\documentclass[10pt,twocolumn,letterpaper]{article}

\usepackage{iccv}
\usepackage{times}
\usepackage{epsfig}
\usepackage{graphicx}
\usepackage{amsmath}
\usepackage{amssymb}
\usepackage{booktabs}
\usepackage{balance}
\usepackage{lipsum}
\usepackage{caption}
\usepackage{algorithm}
\usepackage{algorithmic}
\usepackage{enumerate}
\usepackage{amsmath,amsthm,amssymb}
\usepackage{engord}
\usepackage{bm}
\usepackage{bbm}
\usepackage{amstext}
\usepackage{comment}
\usepackage{color}
\usepackage[x11names,table]{xcolor}
\usepackage{ctable}
\usepackage{booktabs}
\usepackage{multirow}
\usepackage{mathtools}
\usepackage{setspace}
\usepackage{pifont}
\usepackage{dsfont} 
\usepackage[inline]{enumitem}
\usepackage[toc,page]{appendix}
\usepackage{hhline}
\usepackage{wasysym}
\usepackage[accsupp]{axessibility} 

\usepackage[pagebackref=true,breaklinks=true,letterpaper=true,colorlinks,citecolor=citecolor, bookmarks=false]{hyperref}

\usepackage[capitalize]{cleveref}

\iccvfinalcopy 

\def\iccvPaperID{2362} 

\definecolor{Gray}{gray}{0.9}
\definecolor{White}{gray}{1}
\definecolor{DGray}{gray}{0.8}
\definecolor{DDDDGray}{gray}{0.3}
\definecolor{WhiteGray}{rgb}{0.9, 0.9, 0.9}
\definecolor{citecolor}{HTML}{0071bc}
\definecolor{darkred}{rgb}{0.6, 0.1, 0.05}
\definecolor{DeltaColor}{rgb}{0.039,0.73,0.71}
\definecolor{SigmaColor}{rgb}{0.98,0.45,0.0}
\definecolor{AlphaColor}{rgb}{0,0,0.8}
\definecolor{BetaColor}{rgb}{0.8,0,0.8}
\definecolor{GammaColor}{rgb}{0.514,0.34,0.224}
\definecolor{EpsilonColor}{rgb}{0.353,0.725,0.906}
\definecolor{PurpleColor}{HTML}{8B008B}
\definecolor{BadColor}{HTML}{C0392B}
\definecolor{OrangeColor}{rgb}{0.914,0.541,0.0.141}
\definecolor{GreenColor}{rgb}{0.137,0.573,0.565}
\definecolor{RedColor}{rgb}{0.949,0.275, 0.224}
\definecolor{GemBlueColor}{HTML}{0047AB}
\definecolor{LightCyan}{rgb}{0.88,1,1}

\definecolor{bestcolor}{rgb}{1, 0.5, 0.25}
\definecolor{secondbestcolor}{rgb}{1, 0.8, 0.5}

\newcommand{\cate}[1]{\textbf{#1}}
\newcommand{\expref}[1]{\textbf{Exp.~#1.}}
\newcommand{\alphapara}[1]{{\textcolor{red}{-#1}}}

\newcommand{\tabincell}[2]{\begin{tabular}{@{}#1@{}}#2\end{tabular}}

\DeclareMathAlphabet\mathbfcal{OMS}{cmsy}{b}{n}

\newcommand{\colorRef}[1]{\textcolor{red}{#1}}
\crefname{figure}{\colorRef{Fig.}}{\colorRef{Figs.}}
\Crefname{figure}{\colorRef{Figure}}{\colorRef{Figures}}
\crefname{section}{\colorRef{Sec.}}{\colorRef{Secs.}}
\Crefname{section}{\colorRef{Section}}{\colorRef{Sections}}
\crefname{table}{\colorRef{Tab.}}{\colorRef{Tabs.}}
\Crefname{table}{\colorRef{Table}}{\colorRef{Tables}}
\Crefname{equation}{\colorRef{Eq.}}{\colorRef{Eqs.}}
\Crefname{equation}{\colorRef{Equation}}{\colorRef{Equation}}

\newcommand{\projectURL}{\href{https://kailinli.github.io/CHORD}{\tt{https://kailinli.github.io/CHORD}}}

\newcommand\objprior{object-prior\xspace}
\newcommand\supp{Supp. Mat.\xspace}

\newcommand\model{\textsc{Chord}\xspace}
\newcommand\dataset{\textsc{Comic}\xspace}
\newcommand\nimble{NIMBLE\xspace}
\newcommand{\gtpose}{{\color{GemBlueColor}\textbf{\textit{+GT.pose}}}\xspace}
\newcommand{\estpose}{{\color{RedColor}\textbf{\textit{+Est.pose}}}\xspace}

\newcommand{\symgtpose}{{{\color{GemBlueColor}$\CIRCLE$}}\xspace}
\newcommand{\symestpose}{{{\color{RedColor}$\LEFTcircle$}}\xspace}

\newcommand\spc{\ensuremath{\mathcal{SP}}\xspace}
\newcommand\latentcode{\ensuremath{\mathcal{Z}}\xspace}

\newcommand\symbobjprior{\ensuremath{\mathbb{O}}}
\newcommand\symbobjinst{\ensuremath{\mathbf{O}}}

\newcommand\sdfx{\ensuremath{s(\queryx)}\xspace}
\newcommand\sdfxgt{\ensuremath{\widetilde{s}(\queryx)}\xspace}

\newcommand\dhandbefore{\ensuremath{\mathcal{D}_{H}}\xspace}
\newcommand\nhandbefore{\ensuremath{\mathcal{N}_{H}}\xspace}
\newcommand\dobjbefore{\ensuremath{\mathcal{D}_{\symbobjprior}}\xspace}
\newcommand\nobjbefore{\ensuremath{\mathcal{N}_{\symbobjprior}}\xspace}

\newcommand\dobjafter{\ensuremath{\widehat{\mathcal{D}}_{\symbobjinst}}\xspace}
\newcommand\nobjafter{\ensuremath{\widehat{\mathcal{N}}_{\symbobjinst}}\xspace}

\newcommand\dobjgt{\ensuremath{\widetilde{\mathcal{D}}_{\symbobjinst}}\xspace}
\newcommand\nobjgt{\ensuremath{\widetilde{\mathcal{N}}_{\symbobjinst}}\xspace}
\newcommand\dhandgt{\ensuremath{\widetilde{\mathcal{D}}_{H}}\xspace}
\newcommand\nhandgt{\ensuremath{\widetilde{\mathcal{N}}_{H}}\xspace}

\newcommand\feathandnorm{\ensuremath{\mathcal{F}^{\mathcal{N}}_{H}\xspace}}
\newcommand\featobjnorm{\ensuremath{\mathcal{F}^{\mathcal{N}}_{\symbobjinst}\xspace}}
\newcommand\feathanddep{\ensuremath{\mathcal{F}^{\mathcal{D}}_{H}\xspace}}
\newcommand\featobjdep{\ensuremath{\mathcal{F}^{\mathcal{D}}_{\symbobjinst}\xspace}}

\newcommand\shapefeat{\ensuremath{\mathcal{F}_{\rm{S}}}\xspace}
\newcommand\posefeat{\ensuremath{\mathcal{F}_{\rm{P}}}\xspace}
\newcommand\featimg{\ensuremath{\mathcal{F}_{\mathcal{I}}}\xspace}

\newcommand\appearfeat{\ensuremath{\mathcal{F}_{\rm{A}}}\xspace}
\newcommand\transf{\ensuremath{\mathbf{B}}\xspace}
\newcommand\transfinv{\ensuremath{\mathbf{B}^{-1}}\xspace}

\newcommand\rotobjprior{\ensuremath{\mathbf{R}_{\symbobjprior}}\xspace}
\newcommand\tslobjprior{\ensuremath{\mathbf{t}_{\symbobjprior}}\xspace}

\newcommand\handverts{\ensuremath{\mathbf{V}_{H}}\xspace}
\newcommand\joints{\ensuremath{\mathbf{J}}\xspace}

\newcommand\twodnet{\ensuremath{\mathcal{G}_\mathrm{N}}\xspace}
\newcommand\threednet{\ensuremath{\mathcal{G}_\mathrm{S}}\xspace}
\newcommand\image{\ensuremath{\mathcal{I}}\xspace}
\newcommand\queryx{\ensuremath{\mathbf{x}}\xspace}
\newcommand\invqueryx{\ensuremath{\mathbf{x}^{\ominus}}\xspace}

\newcommand\bmapsto{\ensuremath{\boldsymbol{\mapsto}}}
\newcommand\losspix{\ensuremath{\mathcal{L}_\mathrm{pix}}\xspace}
\newcommand\lossvgg{\ensuremath{\mathcal{L}_\mathrm{VGG}}\xspace}

\newcommand{\del}[1]{\xspace}

\newcommand{\cheading}[1]{\noindent\mbox{\textbf{#1}\;}}
\newcommand{\qheading}[1]{\vspace{5pt}\noindent\mbox{\textbf{#1}\;}}
\newcommand{\subqheading}[1]{\vspace{2pt}\mbox{\textit{\textbf{#1}}\;}}
\newcommand{\qparagraph}[1]{\vspace{-1.0em}\paragraph*{#1}}

\newcommand{\xmark}{\ding{55}}%
\newcommand{\greencheck}{{\color{Green4} \checkmark}\xspace}
\newcommand{\bluecheck}{{\color{blue} \checkmark}\xspace}
\newcommand{\redcross}{{\color{red} \xmark}\xspace}

\newcommand{\customfootnotetext}[2]{{
  \renewcommand{\thefootnote}{#1}
  \footnotetext[0]{#2}}}

\newlength\savewidth

\ificcvfinal\fi

\begin{document}

\title{
    \model: Category-level Hand-held Object Reconstruction via \\Shape Deformation
}
\author{
    {
        Kailin Li\textsuperscript{1~$\star$}~~
        Lixin Yang\textsuperscript{1,2~$\star$}~~
        Haoyu Zhen\textsuperscript{1}~~
        Zenan Lin\textsuperscript{3}~~
    }\\{
        Xinyu Zhan\textsuperscript{1}~~
        Licheng Zhong\textsuperscript{1}~~
        Jian Xu\textsuperscript{4}~~
        Kejian Wu\textsuperscript{4}~~
        Cewu Lu\textsuperscript{1,2~$\boldsymbol{\dagger}$}
    } \\
    {
        \small
        {\textsuperscript{1}Shanghai Jiao Tong University}\quad
        {\textsuperscript{2}Shanghai Qi Zhi Institute}\quad
        {\textsuperscript{3}South China University of Technology}\quad
        {\textsuperscript{4}XREAL}
    }
    \\
    {
        \tt\small \textsuperscript{1}\{{kailinli, siriusyang, anye\_zhen, kelvin34501,  zlicheng, lucewu}\}@{sjtu.edu.cn}
    }\\
    {
        \tt\small \textsuperscript{3}auzenanlin@mail.scut.edu.cn\quad 
        \textsuperscript{4}\{jianxu, kejian\}@{nreal.ai}
    }\\
}

\newcommand{\styleFront}{\textbf{\textcolor{frontcolor}{front}}\xspace}
\newcommand{\styleBack}{\textbf{\textcolor{backcolor}{back}}\xspace}
\newcommand{\styleMissingGeometry}{\textbf{\textcolor{sidecolor}{missing geometry}}\xspace}
\newcommand{\styleFaceAndHands}{\textbf{\textcolor{skincolor}{face or hands}}\xspace}

\newcommand{\teaserCaption}{
\textbf{Examples of Hand-held Objects at Category Level.} 
We proposed a new method \model which exploits the categorical shape prior for reconstructing the shape of intra-class objects. In addition, we constructed a new dataset, \dataset, of category-level hand-object interaction. \dataset encompasses a diverse collection of object instances, materials, hand interactions, and viewing directions, as illustrated.
}

\twocolumn[{
    \renewcommand\twocolumn[1][]{#1}
    \maketitle
    \centering
    \vspace{-0.2em}
    \begin{minipage}{1.00\textwidth}
        \centering
        \includegraphics[trim=000mm 000mm 000mm 000mm, clip=False, width=\linewidth]{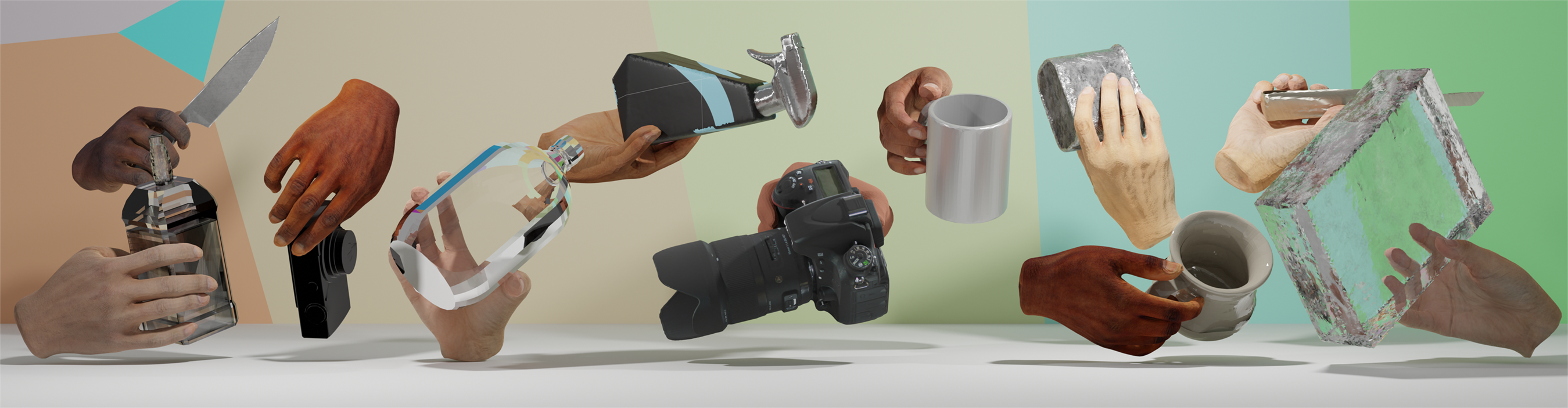}
    \end{minipage}
    \captionsetup{type=figure}
    \captionof{figure}{\teaserCaption}
    \label{fig:teaser}
    \vspace{2em}
}]

\customfootnotetext{$\star$}{
    These authors contributed equally.
}

\customfootnotetext{$\dagger$}{
    Cewu Lu is the corresponding author. He is the member of Qing Yuan Research Institute and MoE Key Lab of Artificial Intelligence, AI Institute, Shanghai Jiao Tong University, China, and Shanghai Qi Zhi institute.
}

\begin{abstract}
    In daily life, humans utilize hands to manipulate objects. 
    Modeling the shape of objects that are manipulated by the hand is essential for AI to comprehend daily tasks and to learn manipulation skills.
    However, 
    previous approaches have encountered difficulties in reconstructing the precise shapes of hand-held objects, primarily owing to a deficiency in prior shape knowledge and inadequate data for training.
    As illustrated, given a particular type of tool, such as a mug, despite its infinite variations in shape and appearance, humans have a limited number of `effective'  modes and poses for its manipulation. This can be attributed to the fact that humans have mastered the shape prior of the `mug' category, and can quickly establish the corresponding relations between different mug instances and the prior, such as where the rim and handle are located.
    In light of this, we propose a new method, \model, for \textbf{C}ategory-level \textbf{H}and-held \textbf{O}bject \textbf{R}econstruction via shape \textbf{D}eformation. \model deforms a categorical shape prior for reconstructing the intra-class objects. 
    To ensure accurate reconstruction, we empower \model with three types of awareness: appearance, shape, and interacting pose.
    In addition, we have constructed a new dataset, \dataset, of category-level hand-object interaction. \dataset contains a rich array of object instances, materials, hand interactions, and viewing directions.
    Extensive evaluation shows that \model outperforms state-of-the-art approaches in both quantitative and qualitative measures. Code, model, and datasets are available at \projectURL    
 \end{abstract}
\vspace{-2em}\section{Introduction}

In daily life, we perform complex tasks by continually manipulating a limited number of simple objects with our hands.  
Understanding the physical nature of hand-object interaction is crucial for AI to comprehend human activities. This requires us to pursue a geometric representation (reconstruction) of the hand-held object, especially for manipulable tools.
Many significant efforts \cite{hasson2019obman,karunratanakul2020grasping,yang2021cpf,yang2022artiboost,Ye2022iHOI,chen2022alignsdf, chen2022tracking} have been made for reconstructing both the hand and hand-held objects. These works, where the hand is represented by a kinematic prior, such as MANO \cite{romero2017embodied}, have achieved high qualities on hand reconstruction via keypoints estimation. 
However, producing the object shape in high quality is more challenging, primarily due to the mutual occlusion, lack of geometrical prior, and insufficient data of variant shapes and appearances.

Most previous works thus resorted to reconstructing objects with known template shape \cite{tekin2019h+o, liu2021semihand,hasson2020hoconsistency,yang2022artiboost, hampali2022kptrasnformer}. This setting is commonly referred to as ``object pose estimation''. However, it would fail on the unseen object instances. 
To address this limitation, some works \cite{hasson2019obman,karunratanakul2020grasping, Ye2022iHOI, chen2022alignsdf}  have attempted to directly regress object-agnostic (and also pose-agnostic) surfaces from images using large-scale synthetic data. However, their results are not robust to varied object shapes and appearances, and often produce ``bubble-like'' shapes and broken geometries. 
These issues can be attributed to either a lack of shape basis \cite{karunratanakul2020grasping,Ye2022iHOI,chen2022alignsdf} or the use of an underlying shape basis with fixed topology and resolution \cite{hasson2019obman, groueix2018atlasnet, kanazawa2018meshrec}.

In this paper, we aim to leverage the best of both worlds: utilizing shape information from a known template 
while also generalizing well to unseen instances.
With this in mind, we propose to first estimate the object pose at category-level, and then resort to a categorical object shape prior (later we call it \textit{\objprior}) to reconstruct the surface of unseen objects.  
We design \textbf{\model}, stand for \textbf{C}ategory-level \textbf{H}and-held \textbf{O}bject \textbf{R}econstruction via shape \textbf{D}eformation.
\model is a deep-learning model that learns to ``deform'' an implicit surface from an explicit \objprior. 
Given the estimated pose of the \objprior, \model takes two steps. 
It first deforms the 2D surface-aware feature maps (\ie normal and depth map) of the object-prior to those of the actual object instance.
From these feature maps, \model then deforms the 3D shape of the object-prior to the actual object instance.

Inspired by \cite{Saito2020pifuhd, Xiu2022ICON, Natsume2018SiCloPeSC}, the first step is achieved by leveraging the design of an image-to-image translation network (\twodnet, \cref{sec:2d_deformation}), 
Specifically, we rendered the normal and depth map of the \objprior in the estimated poses, along with the image, as inputs for \twodnet.
Additionally, we use the rendered depth and normal maps of the estimated MANO hand mesh as extra inputs to help with decoupling the surfaces of the hand and object. 

To perform the second step, we use a point-wise implicit function (IF, denoted as \threednet) \cite{park2019deepsdf} to regress the signed distance  of query points to the surface of object instance. 
The final reconstruction is then obtained as the zero-level set of these query points.
The use of IF enables us to reconstruct objects with fine-grained geometry and arbitrary topology.
We argue that for robustly and accurately reconstructing the hand-held object, it is  necessary for 
\model to possess three different types of awareness: (1) the \textbf{appearance} awareness, (2) the \textbf{shape} awareness of the categorical object-prior, and (3) the \textbf{pose} awareness of the interacting hand.  
To integrate  these awareness, we develop three types of local features (\cref{sec:3d_deformation}\alphapara{A,B,C}). Consequently, \threednet's prediction of signed distance is conditioned on these features.

The final issue is that the current datasets for hand-object interaction (HOI) are not suitable for reconstruction in category-level.
These datasets commonly lack diverse samples within the same category \cite{hampali2020ho3dv2, Chao2021DexYCB, garcia2018FPHAB, cao2020rhoi} or lack real-world human interacting behaviors \cite{hasson2019obman}. To address this limitation, we propose a new dataset, named \dataset, which is built from high-fidelity rendering and targets on category-level hand-object reconstruction (\cref{sec:comic_dataset}). 
This dataset contains a large number of images that depict the interaction between the hand and categorical objects with diverse shapes and appearances (see \cref{fig:teaser,fig:qualitative_comic_test}). Unlike the simulated grasping poses as in GraspIt \cite{graspit}, which do not reflect the intention of human behaviors, the interacting poses in \dataset are based on real-world human demonstrations captured in \cite{Yang2022OakInk}. 

We conduct an extensive evaluation of \model utilizing the \dataset dataset, including qualitative outcomes on several additional HOI datasets (not incorporated for training) as well as unseen, in-the-wild images. 
Our quantitative comparisons verify that \model exceeds the state-of-the-art (SOTA) in performance. Furthermore, our qualitative results illustrate that \model generalizes more effectively to unseen, in-the-wild instances. We summarize our contribution in three-folds:

\begin{itemize}[leftmargin=3.2mm]
    \item We propose the reconstruction of hand-held objects at category-level via our novel model, \model. The model explicitly encapsulates shape from an object-prior, facilitating the deformation of the implicit surface to actual object instances.
    \item Within \model, we incorporate three types of awareness - appearance, shape, and interacting pose - to ensure the accuracy of reconstruction. An extensive ablation study and gain analysis substantiate the theory that performance enhancement corresponds with increased awareness.
    \item We introduce a new dataset for category-level hand-object reconstruction, known as \dataset. This dataset comprises a wealth of high-fidelity images featuring a vast variety of objects and interacting hand poses.
\end{itemize}

\section{Related Work}

\cheading{Hand-held Object Reconstruction.}
Previous studies have utilized RGB or RGB-D data for reconstructing hand-held objects with optimization \cite{pham2017hand, tsoli2018joint}. However, these methods generally only reconstruct a limited number of known objects for which a 3D model is available \cite{garcia2018FPHAB, cao2020rhoi, tekin2019h+o, hampali2020ho3dv2, tzionas2016capturing}, using approaches such as implicit feature fusion \cite{chen2021joint, liu2021semihand, tekin2019h+o}, explicit geometric constraints like contact or collision \cite{brahmbhatt2020contactpose, cao2020rhoi, corona2020ganhand, grady2021contactopt, hamer2010object, zhang2020perceiving}, or physical realism to aid in hand joint reasoning and object reconstruction \cite{pham2017hand, tzionas2016capturing}. These methods commonly assume the availability of the 3D object template during inference, which is a limiting assumption.
Reconstructing hand-held objects without a known template is significantly more challenging as it requires the algorithm to handle various object appearances and shapes while only partially observing them. Hasson \etal \cite{hasson2019obman} utilize a view-centric variant of AtlasNet \cite{groueix2018papier} to handle generic object categories without the need for instance-specific knowledge.
Karunratanakul \etal \cite{karunratanakul2020grasping} characterize each point in 3D space using signed distances to the surface of the hand and object, and combine the hand, object, and contact area in a shared space represented by implicit surfaces. Additionally, works like \cite{Ye2022iHOI,chen2022alignsdf} also focus on reconstructing hand-held objects from images without knowledge of their 3D templates. They use an implicit network to infer the signed distance to parameterize objects and infer the object shape in a normalized hand-centric coordinate space.
In this paper, we focus on what the above approaches miss: reconstructing hand-held objects by using a priori knowledge of shape at the category-level.
	
\qheading{Category-level Object Pose Estimation.}
Category-level object pose estimation involves localizing and estimating the 3D pose of an object within a specific category (e.g., chair, table, or car) without knowledge of the particular instance. 
Sahin \etal \cite{sahin2018category} was the first work to address the problem of 6DoF object pose estimation at the category level. However, its generalization capability across unseen object instances is limited.
One of the significant challenges for category-level pose estimation is the intra-class variation, including appearance and shape. To address this challenge, Wang \etal \cite{Wang2019NOCS} proposed a shared canonical representation, called Normalized Object Coordinate Space (NOCS), that uses the dense image-shape correspondence for estimating target instances under the same category. Tian \etal \cite{tian2020shape} used the NOCS and also explicitly modeled the deformation from a pre-learned categorical shape prior.
Chen \etal \cite{chen2020CASS} modeled canonical shape space (CASS) as a latent space with normalized poses for estimating the  6DoF poses of objects. 
Furthermore, Chen \etal \cite{chen2022category} proposed a Decoupled Rotation Representation that directly learns the two perpendicular vectors of a categorical-specified object, just name a few. 
Mainstream category-level object pose estimation follows the NOCS formulation, \cite{Wang2019NOCS, chen2020CASS, tian2020shape, Fu2022CategoryLevel6O}, which requires the extra depth map as input. As an exception, OLD-Net \cite{fan2022oldnet} eliminates the need for depth by additionally performing depth reconstruction. 
In our task, where the interacting hand is predictive of the object shape and scale, the requirement for a depth map can be replaced by incorporating the hand's pose estimation. 
For category-level object pose estimation, we employ decoupled rotation axes for the sake of simplicity. This representation eliminates the need for dense image-to-shape correspondences such as NOCS, which can be particularly challenging to establish in the presence of severe occlusions between the hand and object.

\section{Method}
\label{sec:method}

Inferring the detailed shape of both the hand and object of known category is a challenging and ill-posed problem: the hand commonly occludes the object, and vice versa, making it necessary for the deep network to decouple the conjunct parts and reconstruct the surfaces separately, based on the observable parts.
\model jointly considers a MANO  hand model and the \objprior to reduce the ambiguities (\cref{sec:preceding_task}).
Specifically,
\model takes an RGB image cropped around the hand, along with the estimated hand (MANO) parameters and the pose of a category-level  \objprior, and outputs the shape reconstruction of the object instance aligned with the input image. 
\model uses two consecutive submodules to this end: (1) pixel-level 2D deformation (\cref{sec:2d_deformation}) and (2) point-level 3D deformation (\cref{sec:3d_deformation}).
After completing these two steps, \model outputs the shape of an unseen object instance in the form of zero-level set of signed distance field (SDF).

\subsection{Preceding Task Model}
\label{sec:preceding_task}
To estimate the MANO hand mesh, we leverage the 3D keypoints prediction \ensuremath{\joints \in \mathbb{R}^{21 \times 3}} as proxy following \cite{zhou2020monocular}, since keypoints are more robust to occlusion. 
The 3D keypoints are then transferred to MANO parameter via an inverse kinematics network (IKNet) \cite{zhou2020monocular}. 
The MANO parameter consist of pose $\bm{\theta} \in \mathbb{R}^{K \times 3}$, and shape $\bm{\beta} \in \mathbb{R}^{10}$, which together are mapped to the MANO hand mesh \ensuremath{\handverts = \mathcal{M}(\bm{\theta}, \bm{\beta}) \in \mathbb{R}^{N_V \times 3}} via a skinning function $\mathcal{M}$ \cite[Eq.(1)]{romero2017embodied}. The $K=16$ and $N_V=778$ are the numbers of joint rotations and vertices of MANO, respectively.
For estimating the rotation of the \objprior from the camera, we use the category-aware Decoupled Rotation representation following \cite{chen2021fs-net}, where the two (or one, depending on the object's symmetrical property) shape-aligned rotation axes $\mathbf{R}_1, \mathbf{R}_2 \in \mathbb{R}^3$ are predicted separately.
These rotation axes are then used to construct the rotation matrix \ensuremath{\rotobjprior \in SO(3)} of the \textbf{\objprior (denoted as \symbobjprior)}.
For estimating the translation of the \objprior relative to hand  \ensuremath{\tslobjprior \in \mathbb{R}^3}, we employ 1-channel volumetric likelihood heatmap following AlignSDF \cite{chen2022alignsdf}.
Importantly, \model is also compatible with other MANO-based mesh recovery models \cite{chen2021camera, lin2021metro,li2023hybrik}, as well as other category-level object pose representation, such as NOCS \cite{Wang2019NOCS}. 
The simple preceding task models used in \model allow us to isolate the benefits of the shape prior guided 2D and 3D deformations. Network details are provided in \supp

\begin{figure}[t]
    \begin{center}
       \includegraphics[width=1.0 \linewidth]{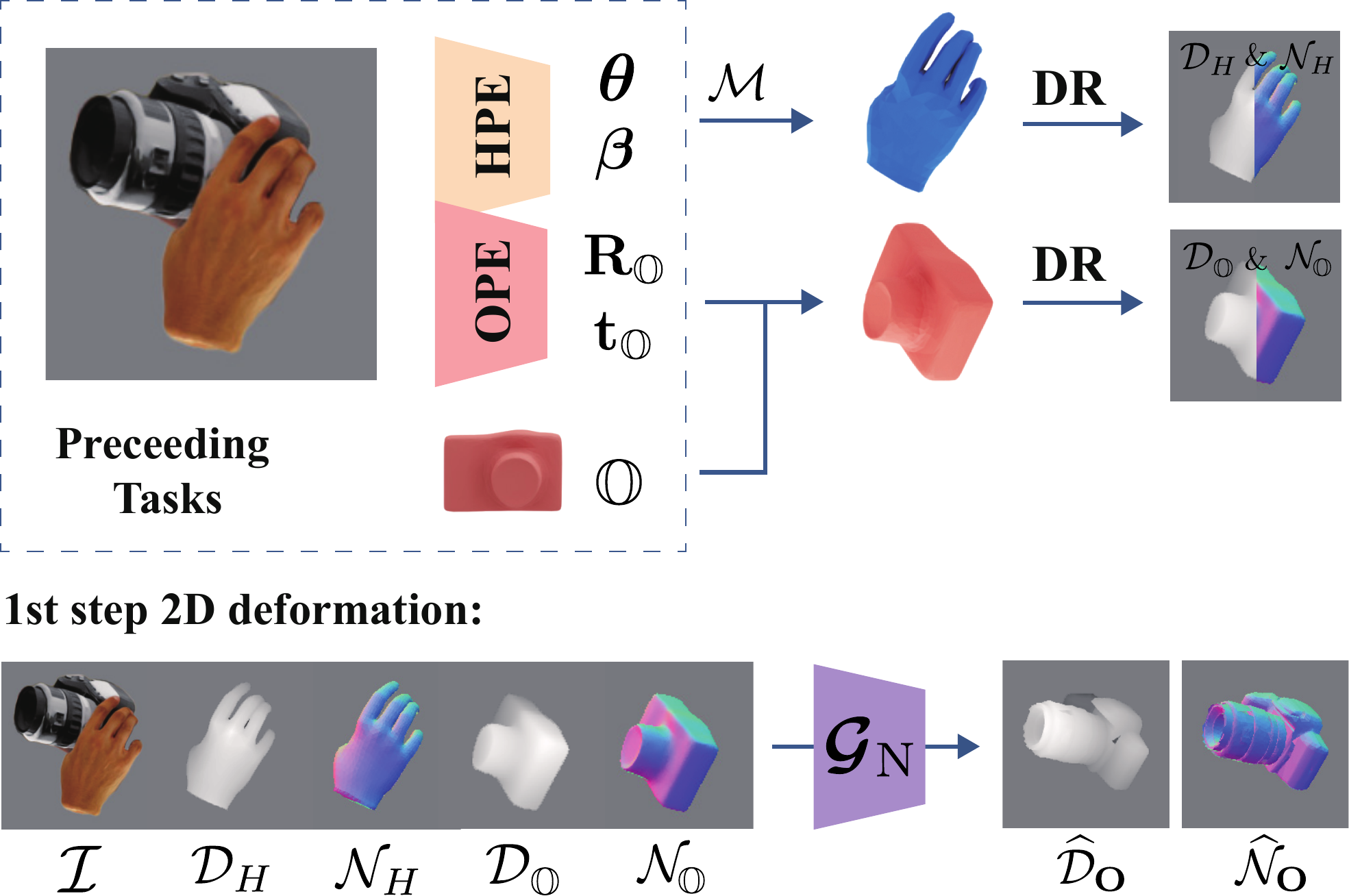}
       \caption{\textbf{Illustration of the first step 2D deformation \twodnet}.
       \textbf{HPE}: hand pose estimation, \textbf{OPE}: category-level \objprior pose estimation, \textbf{DR}: differentiable rendering.}
    \label{fig:2d_deform}
    \end{center}\vspace{-4mm}
\end{figure}

\subsection{Object-prior Guided 2D Deformation}
\label{sec:2d_deformation}

The success of 3D clothed human digitization \cite{Saito2020pifuhd, Xiu2022ICON} has demonstrated two key findings: (1) common convolution neural networks are more adept at inferring 2D feature maps aligned with images than directly estimating detailed 3D surfaces, and (2) the 2D normal map (a common 2D feature map) can serve as a useful guide for 3D surface reconstruction. 
Building upon these observations, in the first step, we utilize an image-to-image translation network \twodnet to predict the surface-aware 2D features (\ie normal and depth map) of the hand and object from the input image.
These image-aligned 2D feature maps are lately used to aid the 3D surface reconstruction, which we introduce in \cref{para:2d_features}\alphapara{A}.

To reduce the ambiguity of the conjunct 3D surfaces, we incorporate the poses of the MANO and object-prior estimated from the preceding task model (\cref{sec:preceding_task}) as additional inputs to \twodnet.
Given the estimated mesh of hand and \objprior,
we use a differentiable renderer in PyTorch3D to generate four 2D feature maps: \dhandbefore and \nhandbefore for the depth and normal maps of hand, and \dobjbefore and \nobjbefore for those of the \objprior .
The separated hand and object depths and normals provide initial guidance for decoupling the 3D surfaces.
Accordingly, the network learns to deform 
the initial feature maps of the \textbf{\objprior (denoted as \symbobjprior)} so as to synchronize them with the \textbf{object instance (denoted as \symbobjinst)} observed in the input image.

Based on these initial 2D feature maps \ensuremath{\{\mathcal{D}_{\star}, \mathcal{N}_{\star}\}}, where \ensuremath{\star \in \{H, \symbobjprior \} }, 
\twodnet predicts the 2D feature maps of the actual object instance post-deformation, denoted as \dobjafter (for the depth map) and \nobjafter (for the normal map).
Formally, the \twodnet represents the following mapping
\begin{align}
    \setlength{\abovedisplayskip}{5pt}
    \setlength{\belowdisplayskip}{5pt}
    \twodnet: (\image, \dhandbefore, \nhandbefore, \dobjbefore, \nobjbefore ) \bmapsto (\dobjafter, \nobjafter).
    \label{eq:2d_deformation}
\end{align}
The input of \twodnet comprises the original RGB image \image, along with the four initial feature maps. 
The loss function for training network \twodnet consists of two terms: (1) a pixel-wise $L$1 discrepancy loss of the two feature maps:  
\ensuremath{\losspix = \sum (| \nobjgt - \nobjafter | + | \dobjgt - \dobjafter |)},  where \nobjgt, \dobjgt are the ground-truth normal and depth map of actual object instance, 
and (2) a perceptual loss \lossvgg \cite{Johnson2016PerceptualLoss} weighted by \ensuremath{\lambda_{\mathrm{VGG}}}.

\subsection{Object-prior Guided 3D Deformation}
\label{sec:3d_deformation}
Given the estimated pose of the \objprior, the MANO hand parameters, and the image-aligned 2D feature maps,
we propose an implicit network \threednet for regressing the 3D surface of the hand-held object. 
To achieve the three types of awareness, 
the network utilizes three types of local features aggregated from three different modalities.
Specifically, for a given query point \queryx, we require its:
(1) 2D pixel-aligned features in normals and depth maps,
(2) 3D shape features interpolated within its nearest region on the \objprior, and
(3) 3D articulated features represented in the local-registered MANO's pose space.
Based on these inputs, the network \threednet deforms the object-prior by predicting the signed distance from \queryx to its closest point on the object instance.

\qparagraph{A. 2D Appearance-aware Feature}\label{para:2d_features}
Given the predicted 2D normal and depth maps of the actual hand and object instance, we extract the local 2D features of the query point \queryx on these feature maps using bilinear interpolation.
The local 2D features consist of four terms: \ensuremath{\feathandnorm(\queryx)}, \ensuremath{\feathanddep(\queryx)}, \ensuremath{\featobjnorm(\queryx)}, and \ensuremath{\featobjdep(\queryx)}. Specifically, \ensuremath{\feathandnorm(\queryx) =  \nhandbefore \big(\pi (\queryx) \big)} represents the normal value at the projection of query point \ensuremath{\pi(\queryx)} on the estimated hand normal map \nhandbefore, and the same principle applies to the remaining terms.
The final appearance-aware features \appearfeat are as:  
\begin{align}
    \setlength{\abovedisplayskip}{5pt}
    \setlength{\belowdisplayskip}{5pt}
    \appearfeat(\queryx) = [\feathandnorm(\queryx),\featobjnorm(\queryx), \feathanddep(\queryx),  \featobjdep(\queryx)].
    \label{eq:affear_feat}
\end{align}

\qparagraph{B. 3D Shape-aware Feature.}\label{para:shape_featurees}
To facilitate the capture of shape-aware features by \model, we incorporate a categorical shape prior for reconstruct different object instances.
Accordingly, instead of learning the implicit surface of different objects separately, as is done in iHOI \cite{Ye2022iHOI} and AlignSDF \cite{chen2022alignsdf}, 
\model learns to regress them from a shared set of latent codes on the \objprior. 
We draw inspiration from the Neural Body \cite{Peng2020NeuralBody}, 
which employs the structured latent codes  anchored on the SMPL body vertices (inner surface) for reconstructing the clothed human body at the outer surface.
Similarly, in \model,  we anchor a set of latent codes \latentcode to the vertices of a mesh-formed \objprior.
These codes are learned alongside the \model model using the images of different object instances within the same category. 
To account for the shape variance between the \objprior and all individual instances, we employ the SparseConvNet \cite{Graham2017sparseconvnet} (denoted as \spc) to diffuse the structured latent codes on the \objprior to its nearby 3D volume space following \cite{shi2020pvrcnn}. 
Then, for a given query point \queryx, we extract its 3D shape feature by trilinear (tri-) interpolation in this diffused volume space: \ensuremath{\spc(\latentcode)}.
Notably, as learning on the structured latent code should  be independent of the pose of the \objprior, we first transform \queryx from world system to \objprior's canonical system using the inverse transformation of \rotobjprior, \tslobjprior before the tri-interpolation. We denote the query point in the object canonical-space as \ensuremath{\invqueryx}, where  
\ensuremath{\invqueryx = \mathtt{inv}(\rotobjprior, \tslobjprior) \cdot \queryx}.
Finally, the shape-aware feature is expressed as:
\begin{align}
    \setlength{\abovedisplayskip}{5pt}
    \setlength{\belowdisplayskip}{5pt}
    \shapefeat(\queryx) = \spc(\latentcode, \invqueryx),
    \label{eq:shape_feat}
\end{align}
where the \ensuremath{\spc(\latentcode, \cdot)} denotes the tri-interpolation on the diffused latent code space.

\qparagraph{C. 3D Pose-aware Feature.}\label{para:pose_features}
To reconstruct an object instance held by hand, it is also  essential for the network to also encapsulate the articulated hand poses.
Therefore, we represent the pose-aware feature as the pose-conditioned value of the query point \queryx. 
The hand pose is described as a set of rigid transformation matrices \ensuremath{ \{\transf \}^{K}_{b=1}}, where each \ensuremath{\transf_b} represents the pose of the \ensuremath{b}-th joint' s local frame in the world system.  
\ensuremath{\transf} can be transformed from MANO pose \ensuremath{\bm{\theta}} using Rodrigues' rotation formula.
As demonstrated in the neural articulated shape approximation literature \cite{deng2020nasa, chen2021snarf}, the query point \queryx is more expressive when represented in the hand's rest-pose system (bone's local frame: \ensuremath{\transfinv_b}). 
In light of this, we design \queryx's pose-conditioned feature \posefeat as its coordinates in the total of \ensuremath{K=16} joints' local frames:
\begin{align}
    \setlength{\abovedisplayskip}{5pt}
    \setlength{\belowdisplayskip}{5pt}
    \posefeat(\queryx) = \{ \transfinv_b(\queryx) \}_{b=1}^{K}.
    \label{eq:pose_feat}
\end{align}

\begin{figure}[!htp]
    \begin{center}
       \includegraphics[width=1.0 \linewidth]{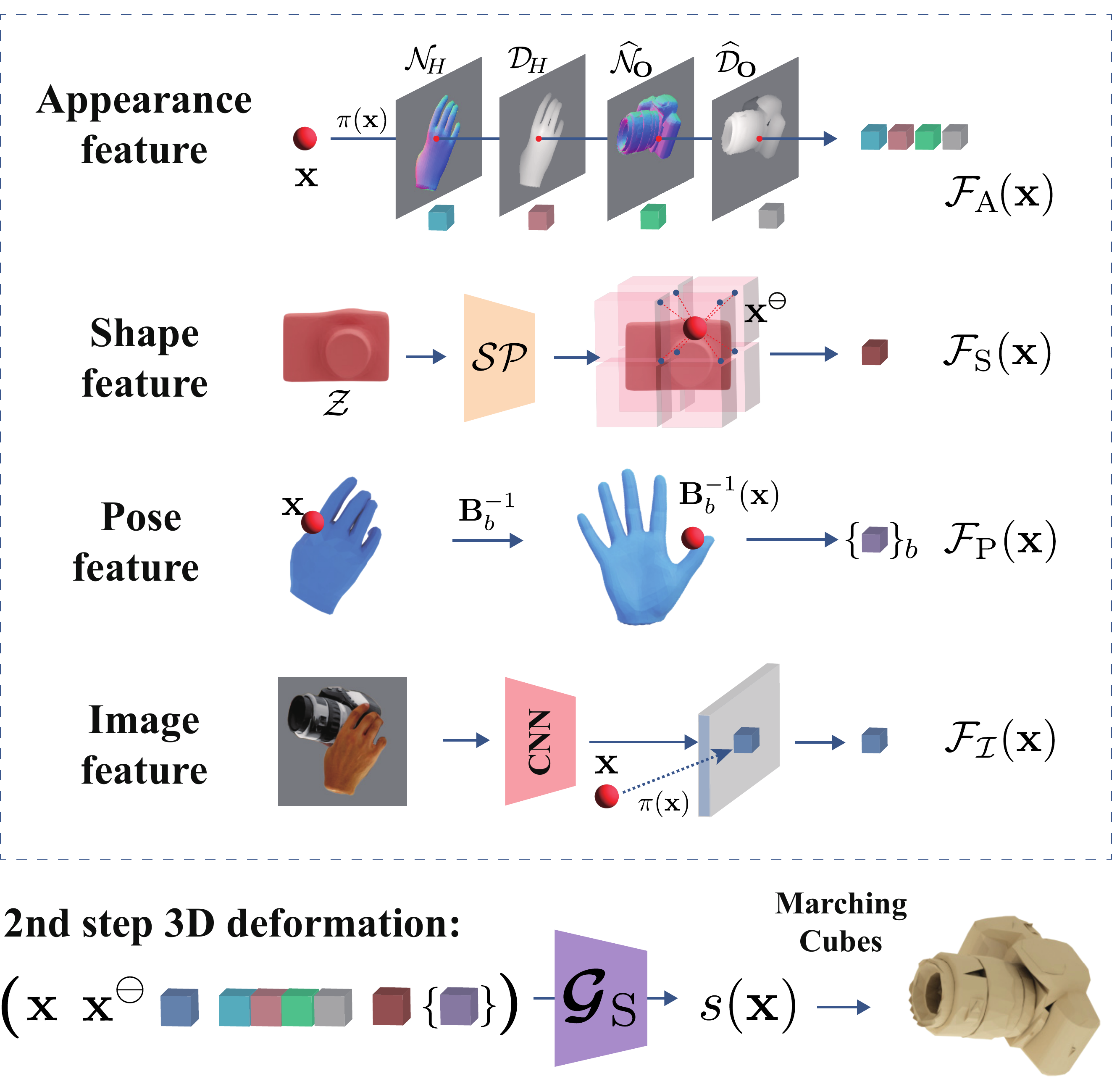}
    \end{center}
    \vspace{-5mm}\caption{\textbf{Illustration of the second step 3D deformation \threednet}, where the three types of awareness (\appearfeat, \shapefeat, \posefeat) are integrated to predict the signed distance of \sdfx.}\vspace{-1em}
    \label{fig:3d_deform}
\end{figure}

\qparagraph{The Implicit Network.}
After collecting the three types of local features, we concatenate them channel-wise to form one input to \threednet. 
In addition, we extract the image feature at the projected coordinate \ensuremath{\pi(\queryx)} on the four feature pyramid layers of the ResNet-50 \cite{resnet} encoder. Inclusion of these features enables the  \threednet to capture a broader range of visual cues at various receptive fields. The 4-layer extracted features are concatenated channel-wise and subsequently mapped to a 16-dimensional feature vector, denoted by \ensuremath{\featimg(\queryx)}.
Beside, the query point's positoins in world \queryx and object-canonical spaces \invqueryx are also used by \threednet as positional encoding, following \cite{chen2022alignsdf}.
The \threednet maps these query points with image feature, appearance-aware, shape-aware, and pose-aware features to the signed distance \sdfx of the object:
\begin{align}
    \setlength{\abovedisplayskip}{5pt}
    \setlength{\belowdisplayskip}{5pt}
    \threednet: (\queryx, \invqueryx, \featimg(\queryx), \appearfeat(\queryx), \shapefeat(\queryx), \posefeat(\queryx)) \bmapsto \sdfx.
    \label{eq:3d_deformation}
\end{align}
We train the \threednet with the $L$1 loss between the predicted \sdfx and the ground-truth \sdfxgt.
When performing inference, we follow the query points sampling strategy as DeepSDF \cite{park2019deepsdf}. The meshes surface are extracted from zero-level set by surface construction algorithm \cite{lorensen1987marchingcubes}.

\subsection{Constructing Object Shape Prior}
\label{sec:object_prior}
\begin{figure}[t]
    \begin{center}
       \includegraphics[width=1.0\linewidth]{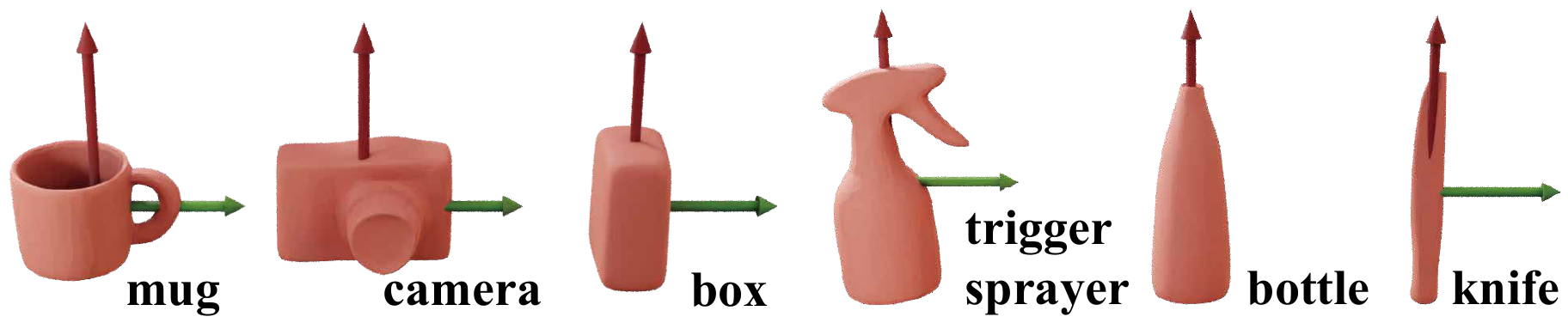}
    \end{center}
    \vspace{-2mm}\caption{\textbf{Illustration of the \objprior.} These prior shapes are extracted using the DIT method, which learns individually across the six categories within the \dataset datset.}\vspace{-4mm}
    \label{fig:shape_prior}
\end{figure}

Given a set of objects of the same category, we explore three different approaches to find a representative object shape prior. 
\textbf{(1) Vanilla voxel mean.} We follow the approach in \cite{Wallace2019FewShotGF} and explicitly construct the \objprior by averaging the voxel representations. 
\textbf{(2) Deep latent mean.} We employ the DeepSDF \cite{park2019deepsdf} network to jointly learn a latent code $c$ of each object and a decoder that generates the implicit surface based on $c$.
In this case, the \objprior is retrieved by forwarding the mean of the learned latent codes to the decoder. In DeepSDF, the latent codes are learned separately, without knowing the common structure among the objects.
\textbf{(3) Deep implicit template (DIT).} 
To obtain a more representative object-prior, we leverage the advanced method: DIT \cite{Zheng2020DeepImpTemp}, which jointly learns the latent code $c$, an instance-specified warping function \ensuremath{\mathcal{W}}, and an instance-irrelevant implicit template \ensuremath{\mathcal{T}}. 
During training, it wraps the implicit template according to different latent codes to model the sign distance of different objects. In this way, the shape of the \objprior is represented as \ensuremath{\mathcal{T}(\cdot)} (see \cref{fig:shape_prior}).
Since the approach (3) establishes strong correspondences across shapes, it achieves the best reconstruction quality on unseen objects (see \cref{para:ablation_prior}\alphapara{C} and \cref{table:prior_ablation}).

\subsection{\textbf{\dataset} Dataset}
\label{sec:comic_dataset}
To address the limitations of existing datasets for category-level hand-held object reconstruction,
we constructed a new dataset, named \textbf{\dataset}, which is built at \textbf{C}ategory-level and contains rich \textbf{O}bjects, \textbf{M}aterials, \textbf{I}nteractions and \textbf{C}amera-views.

We argue that the human's interactions towards a given object category tend to have semantically similar poses, especially for those artificially designed objects. 
Thus, the main focus of \dataset is to increase the diversity of object instances. 
However, recording hand's interaction with real-world objects on a large scale is both time-consuming and expensive.
To overcome this, we have synthesized images based on \textbf{real-world interactions} on \textbf{virtual objects}. 
We use the OakInk \cite{Yang2022OakInk}, a recent dataset of 3D hand-object interactions, as the source.
OakInk features realistic hand interactions with multitude virtual objects, transferred from human demonstrations on a few real-world counterparts.
We select the six categories from OakInk, namely 
\cate{mug}, ~ \cate{camera}, ~\cate{box}, ~\cate{trigger sprayer}, ~\cate{bottle}, and \cate{knife}. The shape priors of these categories are shown in \cref{fig:shape_prior}.
To generate diverse appearances for the hand, 
we convert the current MANO hand poses to those of \nimble \cite{li2022nimble}, and randomly sample an appearance for it. \nimble provides high surface resolution, natural muscle shape, and realistic skin tone.
To add more variation to object appearance, we additionally applied a random material to the object models.
We also randomly pose the camera orientation and set the intrinsics around the object for more diverse viewpoints and perspectives.
Using the Blender \cite{blender} software with a ray tracing engine, we rendered the images in high fidelity.
\dataset comprises 426 K images of 90 K hand-held objects from six frequently used categories.
Several examples are shown in \cref{fig:qualitative_comic_test} and \supp

\section{Experiments and Results}

\subsection{Datasets and Metrics}
When reporting the quantitative and ablation results, 
the \model is trained and tested \textbf{exclusively} on the \dataset dataset.
For generalizing \model to in-the-wild images, we also incorporate several hand$/$hand-object datasets that contain real-world images with \dataset dataset for training. We list these datasets as follows:   
(1) FreiHand \cite{zimmermann2019freihand} and YouTubeHand \cite{kulon2020weakly} for training the hand pose estimation (HPE);
(2) OakInk-Image \cite{Yang2022OakInk} and DexYCB \cite{Chao2021DexYCB} for training HPE, category-level object pose estimation (C-OPE), and our \model;
For qualitative evaluation of in-the-wild images, we additionally capture several `out-of-domain' images with `unseen' object instances. 

To evaluate the quality of our method, we report Chamfer Distance (CD, in $\mathbf{1 \times 10}~ mm^{2}$, CD is measured in terms of squared distance). We randomly sample 30,000 points on the surface of both the ground-truth model and our reconstructed mesh and calculate the average bi-directional point-to-point distances. In addition, we report two physical metrics, \ie penetration depth (PD, in $cm$) and volume (PV, in $cm^3$), to verify the physical plausibility of our method in modeling hand-object interactions, following \cite{Ye2022iHOI, yang2021cpf}. All metrics are reported in camera coordinates system.

\subsection{Evaluation}
For benchmarking \model on \dataset dataset, we report the evaluation results on a total of six categories. 
To explore the effectiveness of different designs in \model, we primarily focus on the mug.
The reasons are in two-fold.
Firstly, the mug represents the only genus-1 object among the six categories, thus exhibiting the most complex geometric topology. Secondly, the mug's thin walls and deep non-convex interior pose challenges for reconstruction.

The quantitative evaluation of the six categories are reported in \cref{tab:cmp_sota} (mug) and \cref{tab:all_cat} (the remaining).

\begin{figure}
    \begin{center}
       \includegraphics[width=1.0 \linewidth]{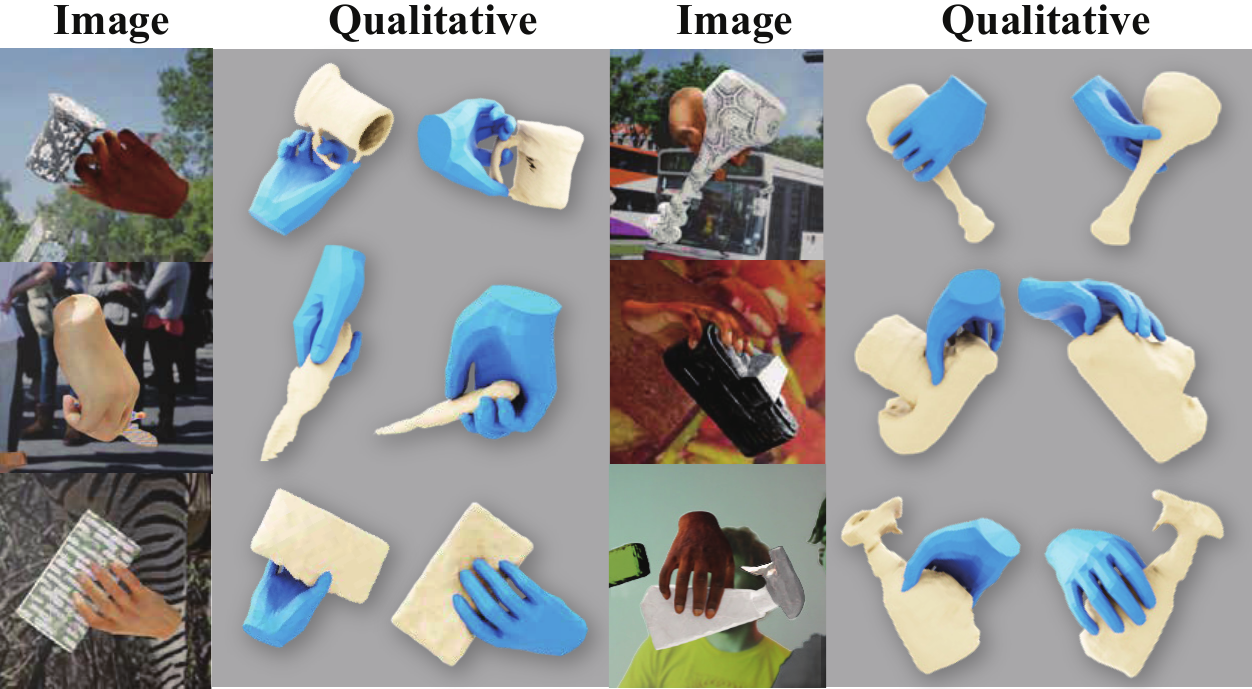}
    \caption{\textbf{Illustration of the images and \model's prediction on \dataset dataset.}}
    \label{fig:qualitative_comic_test}
    \end{center} \vspace{-2em}
\end{figure}

\qparagraph{A. \textbf{\model} -vs- SOTA.} \label{para:vs_sota} 
We compare our method with two recent state-of-the-art hand-held object reconstruction methods: AlignSDF \cite{chen2022alignsdf} and iHOI \cite{Ye2022iHOI}. 
AlignSDF first regresses both the hand pose and object translation, and then performs implicit reconstruction in the aligned pose spaces. iHOI utilizes the estimated hand pose to guide object reconstruction, which corresponds to our \textbf{pose-awareness} in \cref{para:pose_features}\alphapara{C}.
However, neither of these methods is trained at category-level. 
For fairly comparing \model with them, we simulate two category-aware designs, namely, AlignSDF$_C$ and iHOI$_C$.
In short, we incorporate an additional module for \objprior 's pose estimation into these two models.
These two models then predict the signed distance of \queryx, reliant on \objprior 's pose. 
Both networks are trained using our \dataset dataset.
Both the iHOI$_C$ and \model rely on the preceding hand pose estimation. For a fair comparison, we use the same estimated hand pose during the testing phase. 
Technical details are in \supp

To be specific,
when the input pose for the \model are derived from preceding tasks' estimations, 
we refer to this setting as `\estpose'.
Moreover, we conduct an additional set of evaluations utilizing the ground-truth poses of hand and \objprior as inputs to \model. This approach is executed to reveal potential upper-bound performance. 
We refer to this setting as `\gtpose'.

Although \model is not targeting on hand's reconstruction, we empirically find that incorporating the HPE module (recall \cref{sec:preceding_task}) along with \model's second-step, where the two modules share the same ResNet backbone during training, can further improve the HPE results. With this in mind, in SOTA comparison, we also report the mean per joint position error (MPJPE, in $mm$) of hand. 
From \cref{tab:cmp_sota}, we can conclude that our method outperforms the two previous state-of-the-art methods in both reconstruction and physical-related qualities.
We visualize \model's reconstruction results under the \estpose setting in \cref{fig:qualitative_comic_test}.
\begin{table}[t]
    \renewcommand{\arraystretch}{1}
    \begin{center}
    \resizebox{\linewidth}{!}
    {
        \begin{tabular}{l|cc|cc|cc|c}
        \toprule
        \multirow{2}{*}{Methods} & \multicolumn{2}{c|}{CD  $\downarrow$} & \multicolumn{2}{c|}{PD $\downarrow$}& \multicolumn{2}{c|}{PV $\downarrow$}& MPJPE $\downarrow$\\
         & \symestpose & \symgtpose  & \symestpose & \symgtpose & \symestpose & \symgtpose & \symestpose \\
        \midrule
         AlignSDF$_C$ & 8.28 & 3.89 &  0.54 & 0.48 & 3.71 & 3.16 & 6.18 \\
         iHOI$_C$ & 8.14 & 3.80 &  0.54 & 0.48 & 3.61 & 3.03 & 6.20  \\
         \model & \textbf{7.69} & \textbf{3.11} &  \textbf{0.42} & \textbf{0.36} & \textbf{2.79} & \textbf{2.30} &  \textbf{6.08} \\
         \bottomrule
        \end{tabular}
    }
    \caption{ 
        \textbf{\expref{A} Quantitative evaluations on \model \vs SOTAs.}
        $\downarrow$: lower is better; ~\symestpose: \estpose; ~\symgtpose: \gtpose. 
    }
    \label{tab:cmp_sota}
    \end{center} \vspace{-2.0em}
\end{table}

\qparagraph{B. Ablation on Three Awareness.} \label{para:ablation_3_aware} To validate the effectiveness of the three proposed awareness for object reconstruction, we design eight experimental settings. We started with a baseline model that merely used image features as input,  gradually incorporating \textbf{appearance}, \textbf{shape}, and \textbf{pose awareness}, and final reach \model's full design.
\cref{tab:three_awarenesses} shows that the reconstruction performance improves with more awareness types incorporated. The three types of awareness improve the baseline under the predicted pose setting by 36\%, 38\%, and 37\%, respectively. Our full model, \model, achieves the best performance and outperforms the baseline by 40\% and 62\% under the predicted and ground-truth poses settings, respectively.
\begin{table}[b]
    \rowcolors{4}{WhiteGray}{White}
    \renewcommand{\arraystretch}{1}
    \vspace{-0.5em}
    \begin{center}
    \resizebox{0.95\linewidth}{!}
    {
        \setlength{\tabcolsep}{9pt}
        {
        \begin{tabular}{cccc|cc}
        \toprule
        \multirow{2}{*}{\featimg} & \multirow{2}{*}{\appearfeat} & \multirow{2}{*}{\shapefeat} & \multirow{2}{*}{\posefeat} & \multicolumn{2}{c}{Chamfer Distance $\downarrow$} \\
        \hhline{~~~~--}
         &  &  &  & \estpose & \gtpose \\
        \midrule
         \greencheck &\redcross & \redcross &\redcross &  12.88 & 8.16\\
         \greencheck &\greencheck & \redcross &\redcross &  8.25 & 3.78\\
         \greencheck &\redcross & \greencheck &\redcross &  7.99 & 3.45\\
         \greencheck &\redcross & \redcross & \greencheck &  8.14 & 3.80\\
         \greencheck &\greencheck & \greencheck &\redcross &  7.93 & 3.38\\
         \greencheck &\greencheck & \redcross & \greencheck &  7.96 & 3.49\\
         \greencheck &\redcross & \greencheck & \greencheck &  7.94 & 3.29\\
         \greencheck &\greencheck & \greencheck & \greencheck &  \textbf{7.69} & \textbf{3.11}\\
         \bottomrule
        \end{tabular}
        }
    }
    \caption{\expref{B} Gain analysis of the three types of awareness. The \greencheck indicates that the corresponding feature is incorporated by the variant of \model. The \featimg is always used by all variants. }
    \label{tab:three_awarenesses}
    \end{center} \vspace{-2.0em}
\end{table}

\qparagraph{C. Different Object Shape Prior.} \label{para:ablation_prior} In this study, we compare the performance of our \model with different methods of generating \objprior. 
To simplify the experimental setup, we exclusively utilize the \shapefeat feature as input for \threednet, keeping all other variables constant, and report scores under the \estpose setting.
From \cref{table:prior_ablation}, we conclude that \objprior generated by implicit function outperforms that of explicit voxel mean. Among the tested methods, the \objprior from DIT achieves the best reconstruction performance. 
We attribute this improvement to that the learned implicit template has embedded the shape deformation within categories and thus enables the \spc to extract spatial features more effectively.

\input{tables/prior_pose}

\begin{figure*}
    \begin{center}
       \includegraphics[width=1.0 \linewidth]{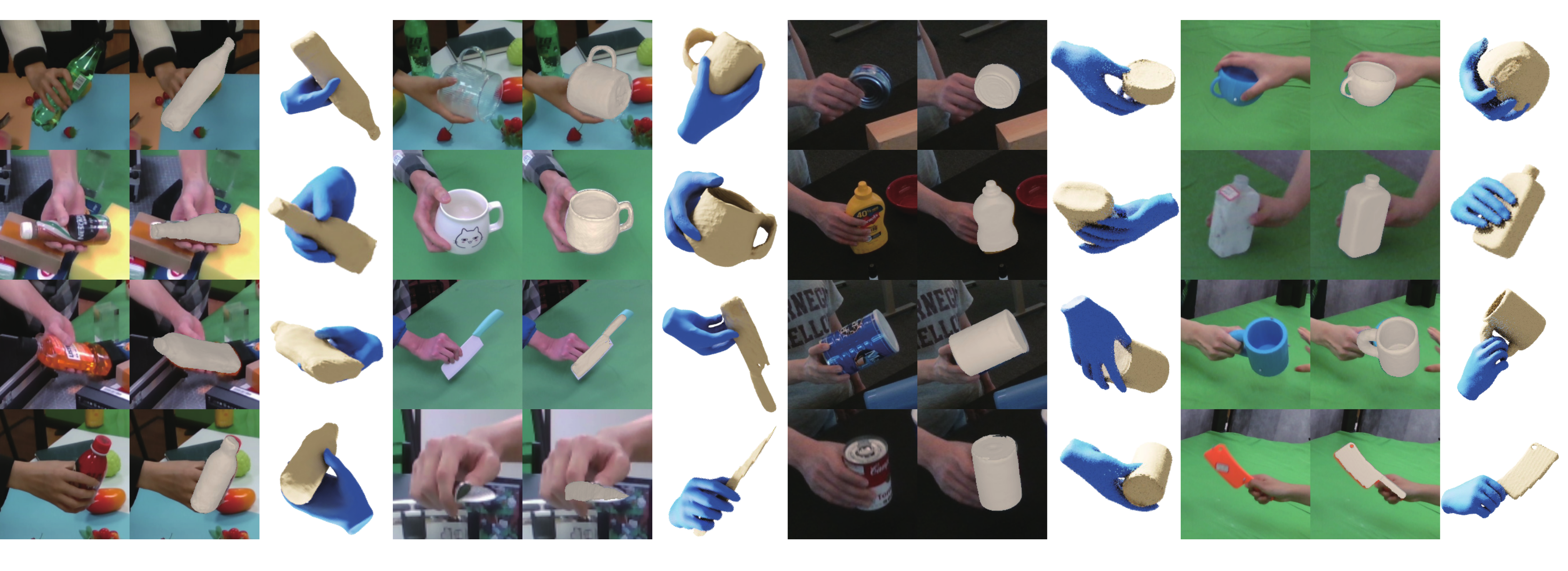}\vspace{-1em}
       \caption{\textbf{Qualitative results of \model's prediction on in-the-wild images.} In the first six columns, the model is tested on images of unseen objects collected in the wild. The results show the reconstruction mesh in the camera view and another free view. The two sets of results on the right demonstrate that CHORD is evaluated on the unseen camera views of DexYCB and OakInk.}     
    \label{fig:qualitative_real_world_img}
    \end{center}\vspace{-1.8em}
\end{figure*}

\qparagraph{D. Different Pose Feature.} \label{para:ablation_poseembedding} 
To explore the way of embedding hand pose into object reconstruction, we propose four experiments under the \estpose setting. (1)  Inspired by the ContactPose \cite{brahmbhatt2020contactpose}, we calculate the distance from each query point \queryx to 21 hand joints, as well as the vector to the nearest point on the hand mesh surface, and its dot product with the surface normal, resulting in \posefeat $\in \mathbb{R}^{23}$. (2) We compute the signed distance of \queryx to the hand mesh using the released neural occupancy model of hand: HALO \cite{karunratanakul2021HALO}, resulting in \posefeat $\in \mathbb{R}^{1}$. (3) Following iHOI, we transfer the \queryx the hand canonical space using the inversion of MANO transformation. Notably, iHOI ignores the \queryx in the wrist (root) aligned system, resulting in \posefeat $\in \mathbb{R}^{45}$. (4) Additionally, we consider the hand's  root transformation. Therefore, our pose-aware feature: \posefeat $\in \mathbb{R}^{48}$. As shown in \cref{table:pose_ablation} the incorporation of information regarding \queryx expressed in the hand canonical space improves the object reconstruction quality.

\begin{table}[b]
\rowcolors{4}{WhiteGray}{White}
    \renewcommand{\arraystretch}{1}
    \vspace{-0.5em}
    \begin{center}
    \resizebox{0.75\linewidth}{!}{
        \setlength{\tabcolsep}{6pt}
        {
        \begin{tabular}{lccccc}
            \toprule
            \nhandgt, \nobjgt  & \redcross & \greencheck & \redcross & \greencheck & \bluecheck\\
            \dhandgt, \dobjgt & \redcross & \redcross & \greencheck & \greencheck & \bluecheck \\
            \midrule
            CD $\downarrow$ & 7.88 & 8.03 & 7.58 & 7.45 & \textbf{7.07} \\
            \bottomrule
        \end{tabular}
        }
    }
    \caption{\expref{E} Ablation study on different appearance features.}
    \label{table:normal_depth_ablation}
    \end{center} \vspace{-2em}
\end{table}

\qparagraph{E. Different Appearance Feature.} \label{para:ablation_normaldepth} 
As presented in \cref{table:normal_depth_ablation}, we conduct an ablation study on the \textbf{appearance awareness} and report score using \estpose setting. Two input forms are explored for normal and depth information: rendering the hand and object together, or rendering them separately. These are indicated by the {\color{Green4}green} \greencheck and {\color{blue}blue} \bluecheck checkmarks, respectively. To minimize the effect of noise and demonstrate the upper bound of \model, we used the rendered ground-truth normal and depth map as input to \appearfeat.
The result in \cref{table:normal_depth_ablation} shows that using separated normal and depth maps as input significantly improves the reconstruction accuracy of the network. This is because the HOI dataset contains a large amount of occlusion, and decoupling the 2D information benefits the subsequent network.

\begin{table}[t]
    \rowcolors{4}{WhiteGray}{White}
    \renewcommand{\arraystretch}{1}
    \begin{center}
    \resizebox{0.9\linewidth}{!}
    {
    
        \setlength{\tabcolsep}{10pt}
        {
        \begin{tabular}{l|ccc}
        \toprule
        \multirow{2}{*}{Catgory} & \multicolumn{3}{c}{Chamfer Distance $\downarrow$, under \estpose} \\
        \hhline{~---}
         & \dataset & OakInk & DexYCB\\
        \midrule
         bottle & 19.86 & 34.37 & 10.24\\
         knife & 69.11 & 32.63 & N/A\\
         camera & 37.85 & 74.97 & N/A \\
         trigger sprayer & 43.07 & 38.81 & N/A\\
         box & 15.50 & N/A & 19.16\\
         \bottomrule
        \end{tabular}
        }
    }
    \caption{\textbf{Results of \model on category-level.} Evaluation on \dataset uses unseen objects, and on DexYCB and OakInk use unseen views. `N/A': the dataset does not contain such category.
    }
    \label{tab:all_cat}
    \end{center}\vspace{-2em}
\end{table}

\vspace{-.2em}
\qparagraph{F. Generalization Ability.} \label{para:in_the_wild} 
The generalization ability of \model is evaluated on two different scenarios. 
The first scenarios relates to existing datasets, whereas the second pertains to real-world `in-the-wild' images. The \model model used in both experiments is trained using the mixture of \dataset, OakInk, and DexYCB datasets.

\subqheading{Existing Dataset.}
Following \cite{Ye2022iHOI,chen2022alignsdf}, we assess \model's generalization ability under the `unseen view' splits using the OakInk and DexYCB datasets.
Specifically, the objects presented during testing are not new to the training set, but are observed from a previously unseen camera viewpoint.
The corresponding scores are listed in \cref{tab:all_cat}. Qualitative results are shown in \cref{fig:qualitative_real_world_img} right.

\subqheading{In-the-wild.}
Second, to unveil potential applications of our \model on real-world in-the-wild scenario, we capture images of hand-held objects in the real-world setting and evaluate \model utilizing these images.
Following iHOI \cite{Ye2022iHOI}, we incorporate hand-object segmentation as an additional input to this process.  
We also estimate the global wrist translation via known camera intrinsic and learned wrist-relative joints' position using \cite[Eq.5]{zhou2020monocular}.
Our model's performance in a real-world context is demonstrated in \cref{fig:qualitative_real_world_img} left, wherein \model exhibits adept reconstructions of previously unseen objects from multiple viewpoints. 
By incorporating prior information, our model achieves remarkable reconstruction results even on objects with complex topology like mug handle. Furthermore, by leveraging our \dataset dataset, the model can generate convincing object meshes, including those for transparent objects. For further details on the real-world settings and additional visualization results, please refer to \supp.

\vspace{-0.2em}
\section{Conclusion}
\vspace{-.1em}
This study introduces a novel method, named \model, for category-level hand-held object reconstruction that overcomes the limitations of prior methods. \model utilizes a pre-trained categorical \objprior and incorporates three types of awareness, namely appearance, shape, and pose, to ensure accurate reconstruction. To address the lack of hand-object data at the category-level, we also introduce a large-scale synthetic dataset, named \dataset, which includes rich object models, realistic materials, diverse hand-object interactions, and camera views. Extensive evaluations demonstrate that \model outperforms SOTA methods in both quantitative and qualitative metrics. The proposed approach has the potential to advance AI's understanding of human activities by accurately modeling the shape of objects that are interacting with hands.
\noindent\rule[0.3ex]{\linewidth}{1.0pt}
{\small
\textbf{Acknowledgments.} This work was supported by the National Key R\&D Program of China (No. 2021ZD0110704), Shanghai Municipal Science and Technology Major Project (2021SHZDZX0102), Shanghai Qi Zhi Institute, and Shanghai Science and Technology Commission (21511101200).
}
\clearpage
{
\small
\balance
\bibliographystyle{configs/ieee_fullname}
\bibliography{egbib}
}
\clearpage

\begin{appendices}
    \balance
    
\section{Implementation Details}\label{sec:supp_impl_details}

To implement \model's 2D deformation networks, \twodnet, we follow the approach described in \cite{Saito2020pifuhd, Xiu2022ICON}, utilizing a pix2pixHD \cite{wang2018pix2pixHD} network design.
To implement \model's 3D deformation networks \threednet,
we select a Multi-Layer Perceptrons (MLP) with five fully connected layers and a single skip connect as the object SDF decoder, which is similar to the decoders used in DeepSDF \cite{park2019deepsdf} and AlignSDF \cite{chen2022alignsdf}. $\threednet$ takes a vector of dimensionality $\mathbb{R}^{94}$ as input. Specifically, $\queryx, \invqueryx \in \mathbb{R}^3$, $\featimg \in \mathbb{R}^{16}$, $\posefeat \in \mathbb{R}^{48}$, $\shapefeat \in \mathbb{R}^{16}$ after dimensional mapping. $\appearfeat$ is a vector obtained from the separately sampled depth and normal maps of the hand and object, \emph{i.e.} $(3 + 1) \times 2 = 8$ dimensions in total.

During inference, we implement a coarse-to-fine reconstruction strategy to secure precise shape reconstructions. The process starts by uniformly sampling $32^3$ query points within a cubic space centered around the object of interest. Subsequently, the signed distance value is computed using the \threednet. 
For spaces yielding initial negative signed distance values, additional $64^3$ points are sampled and their corresponding signed distance values are calculated. 
The final step involves reconstructing the object surface from those signed distances via the Marching Cubes algorithm \cite{lorensen1987marchingcubes}.

\qheading{Quantitative Evaluation.}
To quantitatively evaluate our experiments, 
we first simultaneously train the two preceding tasks (HPE and C-OPE) using the same CNN backbone for 100 epochs.
Then, we train the \model's first-step network \twodnet for 100 epochs. 
We perturb the poses of the MANO \cite{romero2017embodied} and object from the ground-truth data, and use these perturbed poses to generate two meshes: one is the hand mesh obtained by MANO's skinning function \cite{romero2017embodied}, and the other is the \objprior in the perturbed pose. We obtain four 2D feature maps via a differentiable renderer, which serves as the input for \twodnet. We set the weight of the perceptual VGG loss \cite{Johnson2016PerceptualLoss} $\lambda_{\text{VGG}}$ to 0.5.
Finally, we train the \model's second-step network, \threednet, for 100 epochs. 
During training, we use the ground-truths with perturbation as inputs for \threednet, while during testing, we gather inputs from the outputs of the preceding tasks.

We use ResNet34 \cite{resnet} as the backbone across all experiments. Our codebase is implemented in PyTorch. 
During training, we set the batch size to 32, the learning rate to $1\times 10^{-4}$ and decays to $1\times10^{-5}$ after 70 epochs.

\qheading{In-the-Wild Generalization.}
In the experiment of in-the-wild generalization, we train \model's two preceding tasks (HPE and C-OPE) separately. 
(1) To perform the hand pose estimation (HPE), we begin by predicting the hand's 3D joints using the Integral Pose Network \cite{sun2018integral}. We then map these 3D joints to MANO's rotations and shape parameters through the Inverse Kinematics Network (IKNet) \cite{zhou2020monocular}. (2) For category-level \objprior pose estimation (C-OPE), we simultaneously regress the \objprior's Decoupled Rotation axes (introduced in \cite{chen2021fs-net}) and center translation (as in \cite{chen2022alignsdf}). These two tasks use two non-shared ResNet34 backbones.   
The datasets used to train the HPE model include FreiHAND \cite{zimmermann2019freihand}, YouTube3D \cite{kulon2020weakly}, OakInk \cite{Yang2022OakInk}, and DexYCB \cite{Chao2021DexYCB}. Likewise, we train the C-OPE model using \dataset dataset (our own), as well as OakInk and DexYCB. 

Similarly, when performing the in-the-wild reconstruction, we utilize the mixture of \dataset, OakInk and DexYCB dataset to train \model's two steps deformation networks (\twodnet and \threednet).

The correspondence between the categories in our \dataset dataset and the objects in the YCB dataset (used by DexYCB) is as follows:
\begin{table}[h]
    \begin{center}
            \resizebox{1.0\linewidth}{!}
            {
            \makeatletter\def\@captype{table}\makeatother
            \setlength{\tabcolsep}{1mm}{
                \setlength{\aboverulesep}{0pt}
                \setlength{\belowrulesep}{0pt}
                \rowcolors{2}{WhiteGray}{White}
                \setlength{\tabcolsep}{10pt}{
                    \begin{tabular}{l|c}
                    \toprule
                    Category & YCB Object (used in DexYCB) \\
                    \midrule
                     Bottle & \tabincell{c}{ 002\_master\_chef\_can, 005\_tomato\_soup\_can, \\ 006\_mustard\_bottle, 007\_tuna\_fish\_can, \\ 021\_bleach\_cleanser} \\
                     Box &  \tabincell{c}{ 003\_cracker\_box, 004\_sugar\_box, 008\_pudding\_box, \\010\_potted\_meat\_can , 009\_gelatin\_box, \\ 036\_wood\_block, 061\_foam\_brick} \\
                     Mug & 025\_mug  \\
                    \bottomrule
                    \end{tabular}
                }}
            }
            \caption{Object Category of YCB Objects in DexYCB.} 
            \label{table:obj_cat}
    \end{center}\vspace{-2.5em}
\end{table}

\section{Experiment Setting Details}\label{sec:exp_setting}

\subsection{\textbf{\model} -vs- AlignSDF$_C$ and iHOI$_C$}
Comparing our results directly with the original AlignSDF \cite{chen2022alignsdf} and iHOI \cite{Ye2022iHOI} is unfair because they are not trained in a category-level setting. Therefore, we retrain both networks on our \dataset dataset. Additionally, since our model relies on the \objprior pose, we integrate the predicted pose \rotobjprior and \tslobjprior into AlignSDF and iHOI. Using \rotobjprior and \tslobjprior, we transfer the query point \queryx into the object prior canonical space, denoted by \ensuremath{\invqueryx = \mathtt{inv}(\rotobjprior, \tslobjprior) \cdot \queryx}. We then used \invqueryx as input for both AlignSDF and iHOI networks:
\begin{align}
    \text{AlignSDF}_C: & (\queryx, \invqueryx, \featimg(\queryx)) \bmapsto \sdfx. \\
    \text{iHOI}_C: & (\queryx, \invqueryx, \featimg(\queryx), \posefeat(\queryx)) \bmapsto \sdfx.
    \label{eq:align_ihoi_deformation}
\end{align}
Notably, we only consider the network branch for object reconstruction.

\subsection{Pose Feature of HALO network}
In the main paper \textcolor{red}{Sec 4.2-D}: experiment settings \#2, we utilized the output of the HALO \cite{karunratanakul2021HALO} network as the  \posefeat in the ablation study. Specifically, HALO takes the input of the position of 21 hand joints and a query point \queryx, and generates an SDF value $\in \mathbb{R}^1$ which indicates the directed distance from \queryx to the hand surface. We experimentally observe that incorporating the hand SDF as input assists the \model in avoiding surface intersections with the hand while reconstructing the object mesh. However, the accuracy of object reconstruction slightly reduces. 

\subsection{Inputs of the Appearance Feature}
\begin{figure}[h]
    \begin{center}
       \includegraphics[width=0.7 \linewidth]{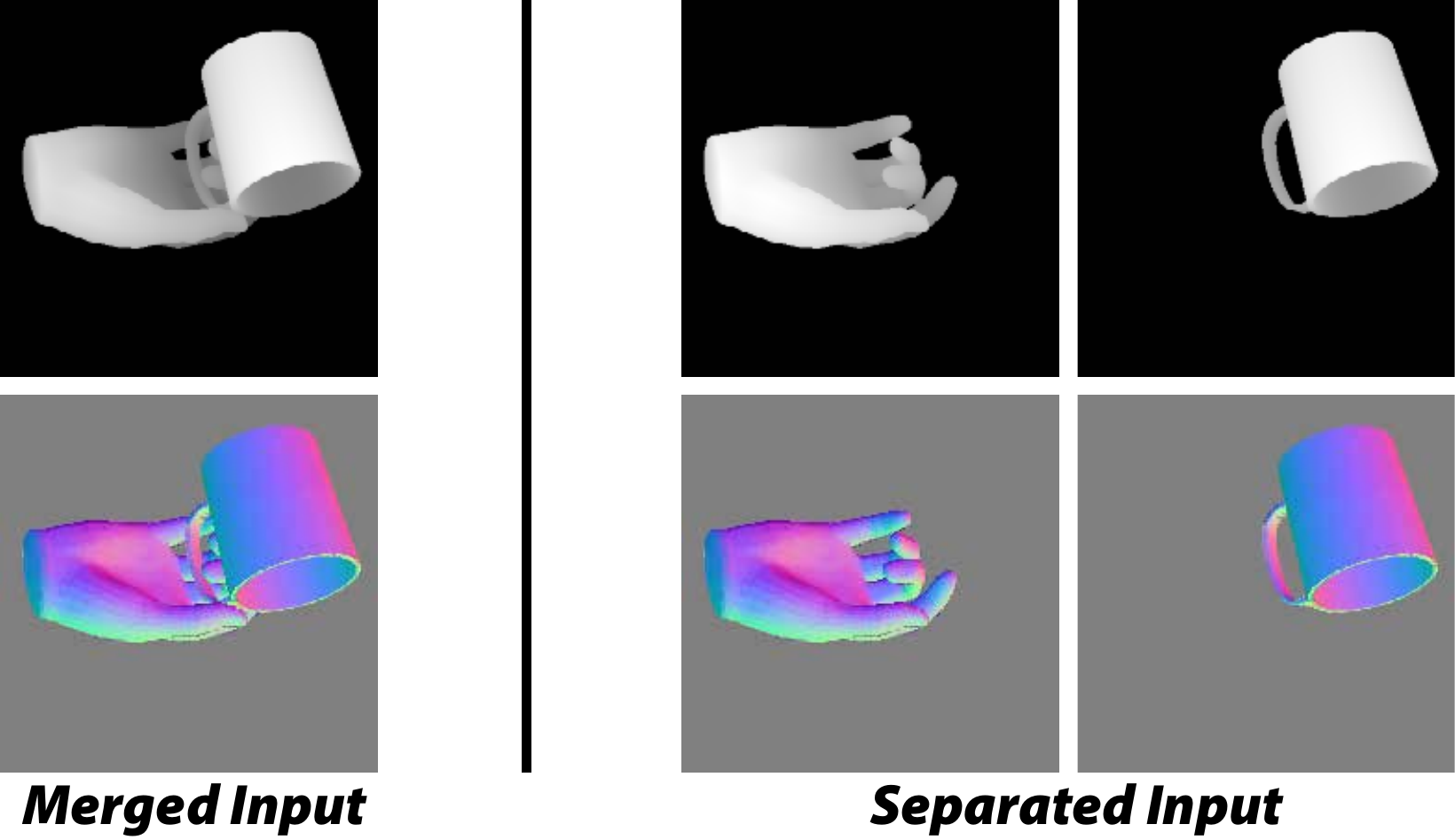}
    \caption{Ablation study on the inputs of \appearfeat.}
    \label{fig:qualitative_supp_normal}
    \end{center} \vspace{-2em}
\end{figure}

In the main paper \textcolor{red}{Sec 4.2-E}, we conduct an ablation study on the normal and depth maps. As shown in \cref{fig:qualitative_supp_normal},  we use Blender to render ground truth maps for minimizing the impact of noise. We train our \model on both the merged hand-object feature map (denoted by the {\color{Green4} green check \greencheck} in Table 5 of the main paper) and the separate hand and object feature maps (denoted by the {\color{blue}blue check \bluecheck}). The results indicate a significant improvement in network performance when using the separate hand and object maps, which help mitigate the occlusion effect between the hand and object.

\section{More Qualitative Evaluation}\label{sec:more_qualitative}

\subsection{\textbf{\model}'s Generalization Ability}
\begin{figure*}[h]
    \begin{center}
        \captionsetup{width=.95\linewidth}
        \includegraphics[width=1.0 \textwidth]{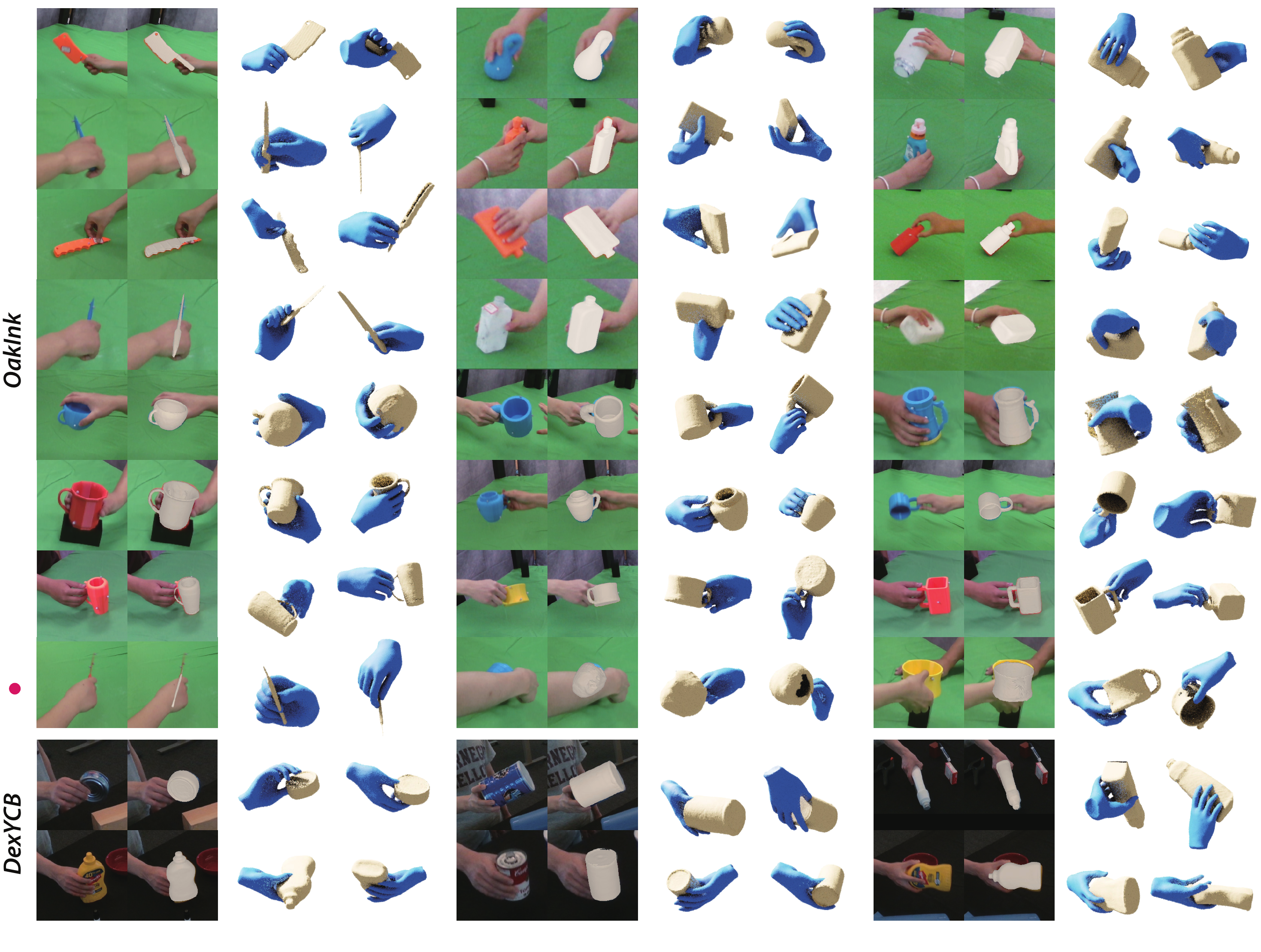}\vspace{-1.5em}
        \caption{\model's results on the OakInk and DexYCB datasets under the \textbf{\color{blue}SO-SD-UC} setting. The line marked by a red circle {\color[RGB]{212,20,90}$\CIRCLE$} indicates the failure cases. \model fails on the objects of extreme thin-wall and severe occlusions.}\vspace{0.5em}
        \label{fig:supp_unseen_view}
        
        \includegraphics[width=1.0 \textwidth]{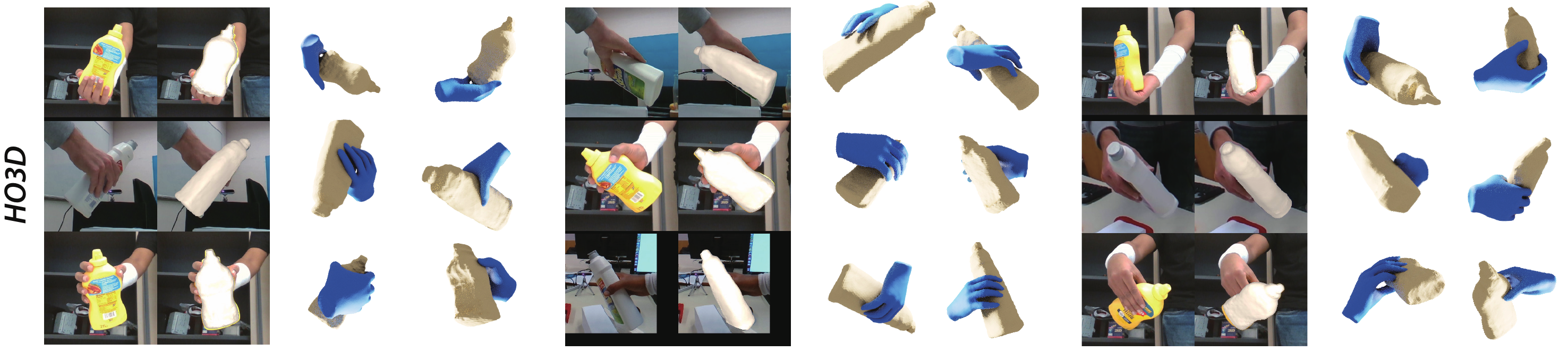}\vspace{-0.5em}
        \caption{The \model's results on the HO3D dataset under the \textbf{\color{GreenColor}SO-UD-UC} setting.}\vspace{0.5em}
        \label{fig:supp_ho3d}

        \includegraphics[width=1.0 \textwidth]{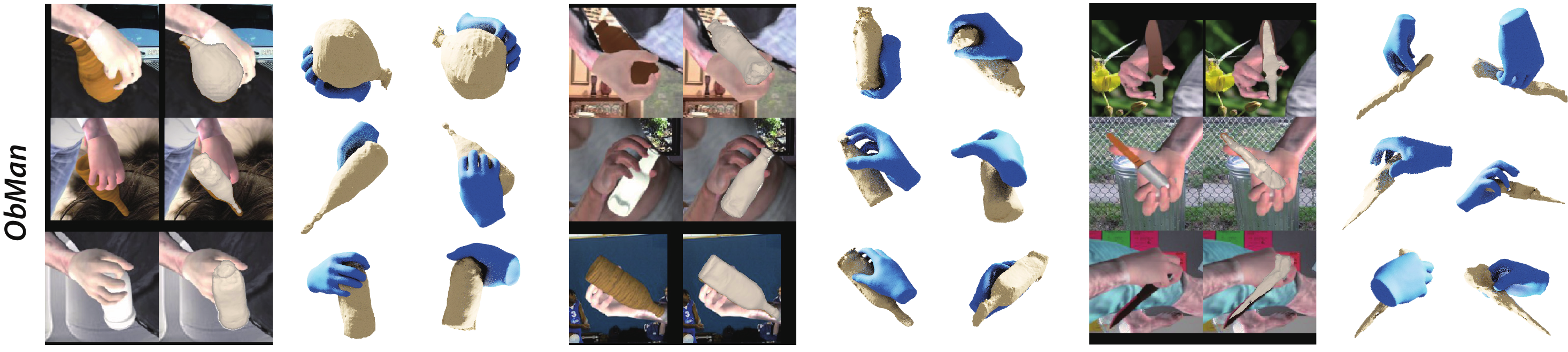}\vspace{-0.5em}
        \caption{The \model's results on the synthetic dataset, ObMan, under the \textbf{\color{OrangeColor}UO-UD-UC} setting.}
        \label{fig:supp_obman}
    \end{center}
\end{figure*}

\begin{figure*}[h]
    \begin{center}
        \captionsetup{width=.95\linewidth}
       \includegraphics[width=1.0 \textwidth]{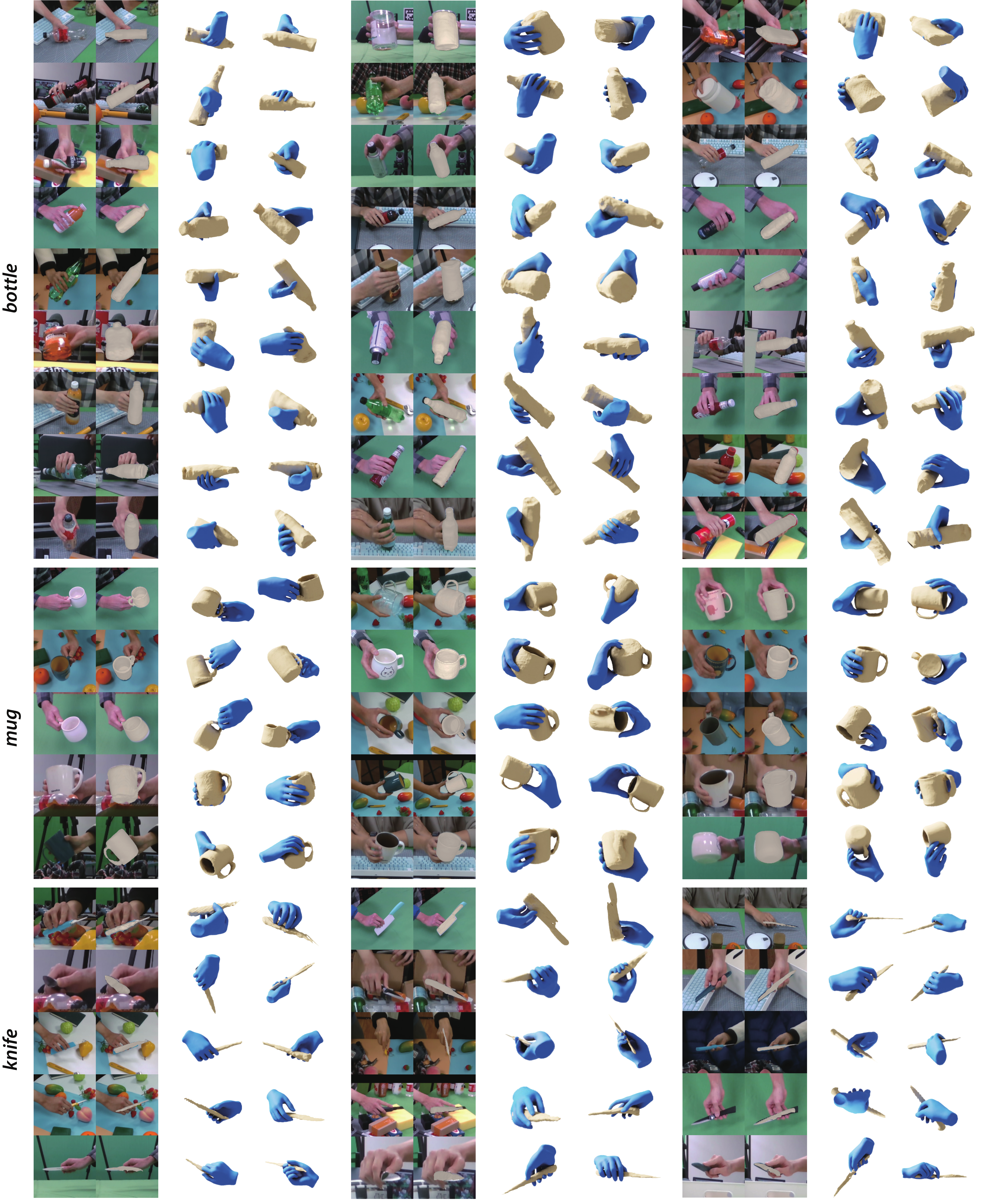}
    \caption{More examples of \model's performance on the in-the wild (\textbf{\color{OrangeColor}UO-UD-UC}) images. The first rows of bottle and mug categories show that our method is capable of accurately reconstructing transparent objects.}
    \label{fig:supp_wild}
    \end{center} \vspace{-2em}
\end{figure*}

We explore the generalization ability of \model under three different settings:

\begin{enumerate}[label={\textbf{(\arabic*)}}]
    \item \textbf{Seen Object, Seen Domain, Unseen Camera-views ({\color{blue}SO-SD-UC})}, where the objects used during testing exist in the training split from the same dataset (same domain), but observed from unseen camera viewpoint.  Evaluation on OakInk and DexYCB testing sets are under this setting (\cref{fig:supp_unseen_view}). 
    \item \textbf{Seen Object, Unseen Domain, Unseen Camera-views ({\color{GreenColor}SO-UD-UC})}, where the objects used during testing exist in the training split of a different dataset (unseen domain) and are observed by unseen camera viewpoint. Evaluation on HO3D dataset is under this setting (\cref{fig:supp_ho3d}). 
    \item \textbf{Unseen Object, Unseen Domain, Unseen Camera-views ({\color{OrangeColor}UO-UD-UC})}, which is also referred to as the \textbf{zero-shot} by iHOI \cite{Ye2022iHOI}.  Evaluation on ObMan \cite{hasson2019obman} and the in-the-wild testing set are under this setting (\cref{fig:supp_obman} and \cref{fig:supp_wild}).
\end{enumerate}

\subsection{Compare with the previous SOTA}
\begin{figure*}[h]
    \begin{center}
       \includegraphics[width=1.0 \textwidth]{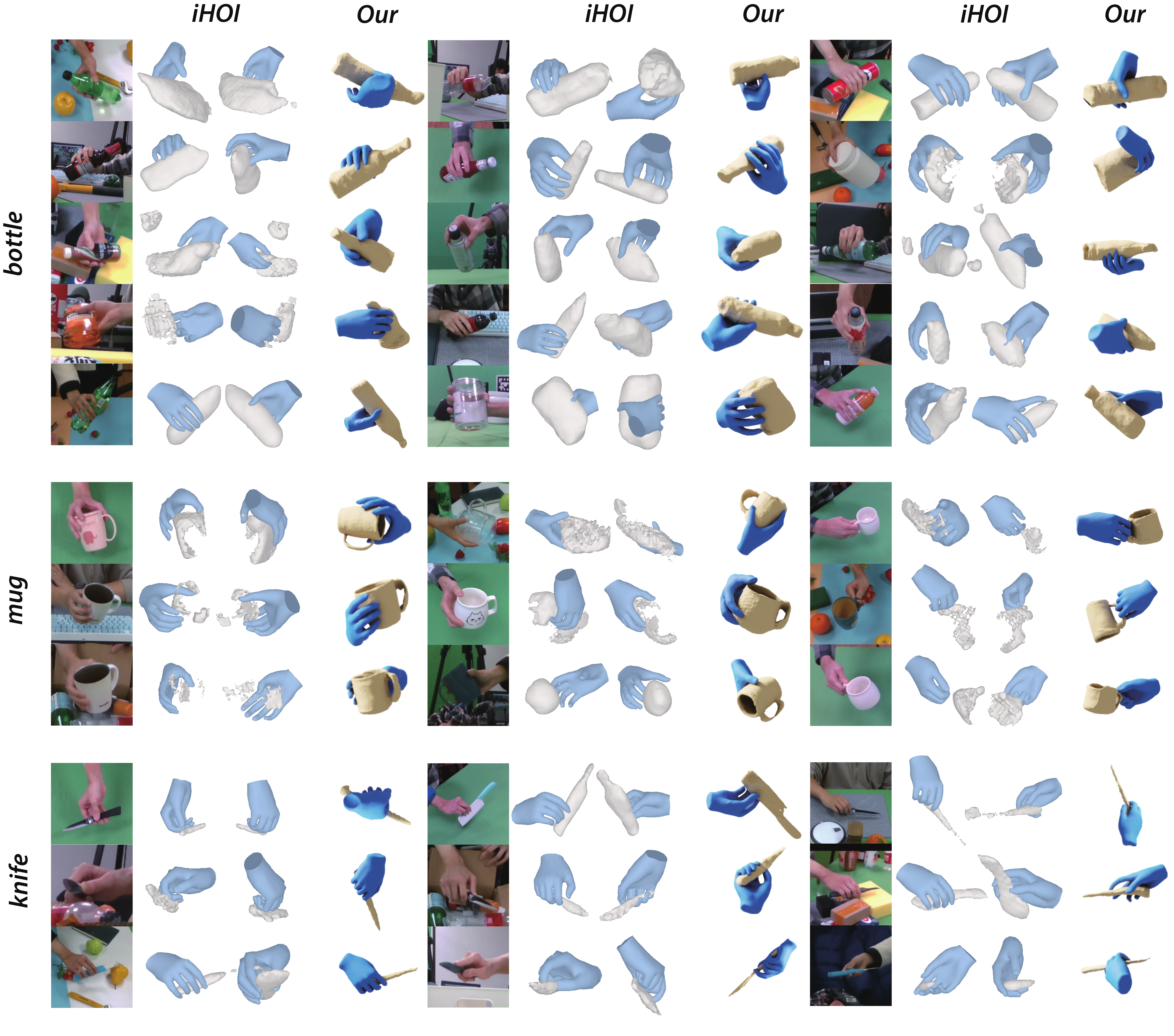}
    \caption{Comparison between our \model and the previous state-of-the-art, iHOI \cite{Ye2022iHOI}.}
    \label{fig:supp_ihoi_cmp}
    \end{center} \vspace{-2em}
\end{figure*}

In \cref{fig:supp_ihoi_cmp}, we qualitatively compare our \model with iHOI \cite{Ye2022iHOI}. We use the iHOI's officially released model and code\footnote{\href{https://github.com/JudyYe/ihoi}{\tt{github.com/JudyYe/ihoi}}}. For a fair comparison, we pass the same predicted hand pose and hand-object mask to the iHOI and our \model model.

\subsection{Examples in \textbf{\dataset} dataset.}
\begin{figure*}[h]
    \begin{center}
       \includegraphics[width=0.9 \textwidth]{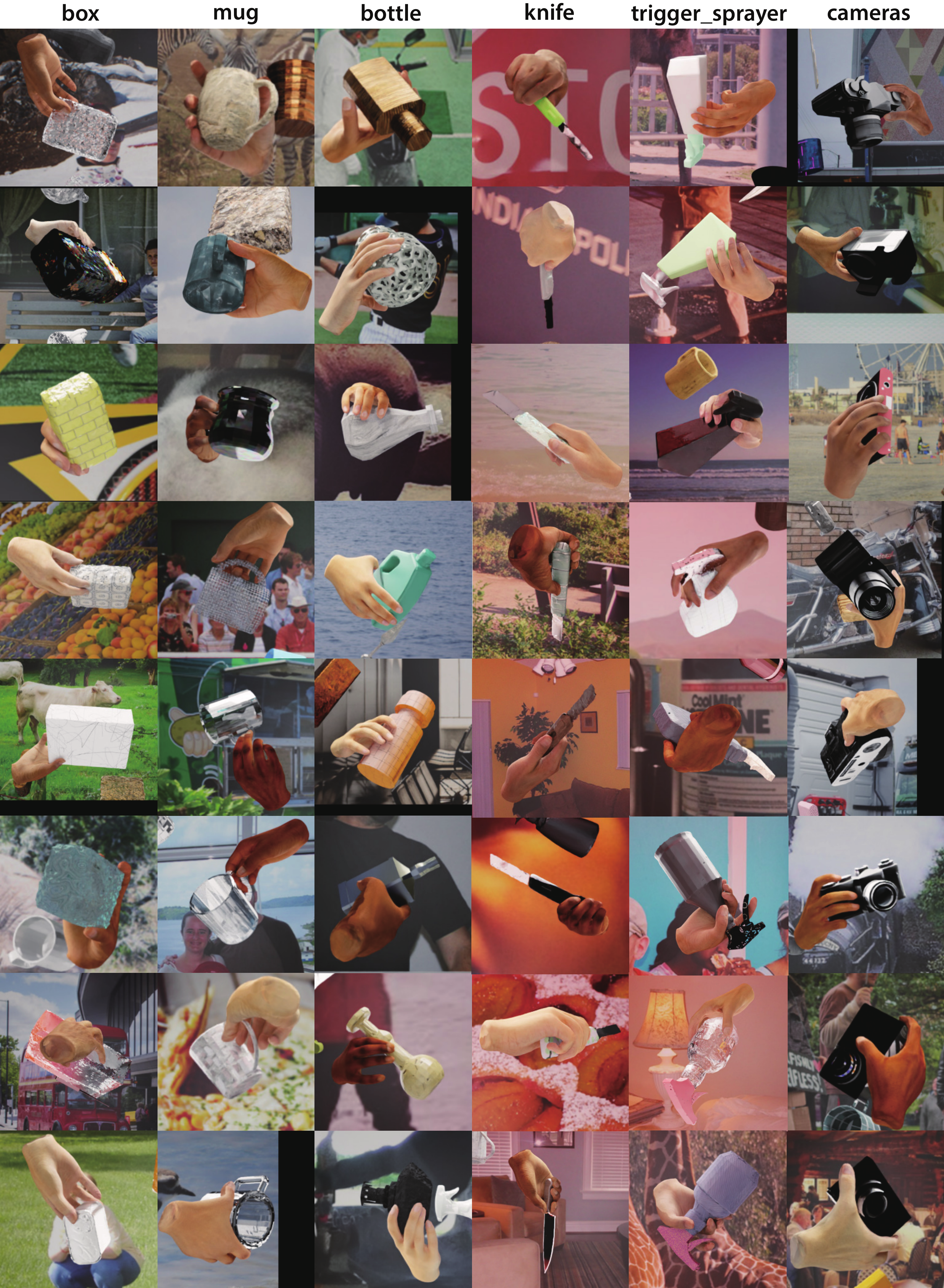}
    \caption{Examples of the six categories in our \dataset dataset.}
    \label{fig:supp_comic}
    \end{center} \vspace{-2em}
\end{figure*}

We show more examples of \dataset in \cref{fig:supp_comic}.

\end{appendices}

\end{document}